\documentclass[lettersize,journal]{IEEEtran}
\usepackage{times}
\usepackage{bm}

\usepackage[numbers]{natbib}
\usepackage{multicol}
\usepackage[bookmarks=true]{hyperref}
\usepackage{algorithm}
\usepackage{algorithmic}
\usepackage{amsmath}
\usepackage{xcolor}
\usepackage{booktabs,multirow}
\usepackage[table]{xcolor}
\usepackage{subcaption}
\usepackage{pifont}

\usepackage{graphicx}
\definecolor{commentblue}{RGB}{0,100,180}
\usepackage{amssymb}

\newcommand{\cmark}{\textcolor{green!60!black}{\ding{51}}}
\newcommand{\xmark}{\textcolor{red}{\ding{55}}}
\newtheorem{theorem}{Theorem}
\newtheorem{lemma}{Lemma}

\begin{document}

\title{Tube Diffusion Policy: Reactive Visual-Tactile Policy Learning for Contact-rich Manipulation}

\author{Teng Xue$^{1, 2, 3}$, Alberto Rigo$^{1}$, Bingjian Huang$^{1, 4}$, Jiayi Shen$^{1}$, Zhengtong Xu$^{1, 5}$, \\ Nick Colonnese$^{1}$, Amirhossein H. Memar$^{1}$
\thanks{$^1$Meta Reality Labs Research, $^2$Idiap Research Institute, $^3$École Polytechnique Fédérale de Lausanne (EPFL), $^4$University of Toronto, and $^5$Purdue University.}
\thanks{This work was conducted during internships at Meta Reality Labs.}
\thanks{Project website: \url{https://schortenger.github.io/tube-diffusion-policy/}.}
}

\let\oldtwocolumn\twocolumn
\renewcommand\twocolumn[1][]{%
    \oldtwocolumn[{#1
    \begin{center}
        \includegraphics[width=0.8\textwidth]{images/overview.png}
        \captionof{figure}{\textbf{Overview of the full pipeline.}
        We first collect visual-tactile human demonstrations via teleoperation in both simulation and on a physical robot, using a robotic arm equipped with a dexterous hand. Based on these demonstrations, Tube Diffusion Policy (TDP) learns a hybrid action velocity field that combines diffusion-based action generation with streaming conditional feedback flows, forming an action tube that constrains the generated trajectory around the demonstration manifold. This design enables reactive control at every timestep, allowing rapid adaptation to contact uncertainty and external disturbances. Moreover, stepwise correction reduces the required denoising steps and significantly lowers inference latency. We evaluate TDP on three virtual and two physical visual–tactile dexterous manipulation tasks, demonstrating strong performance and consistent gains over state-of-the-art baselines.}
        \label{fig:firstpage}
    \end{center}
    \vspace{1em}
    }]
}
\maketitle

\begin{abstract}

Contact-rich manipulation is central to many everyday human activities, requiring continuous adaptation to contact uncertainty and external disturbances through multi-modal perception, particularly vision and tactile feedback. While imitation learning has shown strong potential for learning complex manipulation behaviors, most existing approaches rely on action chunking, which fundamentally limits their ability to react to unforeseen observations during execution. This limitation becomes especially critical in contact-rich scenarios, where physical uncertainty and high-frequency tactile feedback demand rapid, reactive control. To address this challenge, we propose Tube Diffusion Policy (TDP), a novel reactive visual–tactile policy learning framework that bridges diffusion-based imitation learning with tube-based feedback control. By leveraging the expressive power of generative models, TDP learns an observation-conditioned feedback flow around nominal action chunks, forming an action tube that enables fast and adaptive reactions during execution.
We evaluate TDP on the widely used Push-T benchmark and three additional challenging visual–tactile dexterous manipulation tasks. Across all benchmarks, TDP consistently outperforms state-of-the-art imitation learning baselines. Two real-world experiments further validate its robust reactivity under contact uncertainty and external disturbances. Moreover, the step-wise correction mechanism enabled by action tube significantly reduces the required denoising steps, making TDP well suited for real-time, high-frequency feedback control in contact-rich manipulation.
\end{abstract}




\section{Introduction}

Contact-rich manipulation plays a fundamental role in human daily activities and inherently involves rich physical interactions among robots, objects, and the environment. Humans are able to perform such tasks with highly reactive behaviors that adapt instantaneously to contact uncertainty and multi-modal observations. However, to achieve similar capabilities on robots, there are still two open challenges. First, contact uncertainty arises from unknown or varying object geometry, inertial properties, and frictional characteristics, leading to highly nonlinear dynamics \cite{Xue23ICRA, Xue25IJRR}. Second, multi-modal perception—particularly vision–tactile sensing—introduces heterogeneous observation modalities, where tactile signals provide contact-related information at high frequency and demand rapid control responses \cite{xue2025reactive}. Together, these challenges highlight the necessity of reactive control policies that can precisely model complicated behaviors while quickly adapt to new observations in real time.

Recently, imitation learning based on generative models has emerged as a dominant paradigm for learning robot control policies, such as Diffusion Policy \cite{chi2023diffusionpolicy} and Flow Matching Policy \cite{zhang2024affordance}. These approaches excel at capturing the multi-modal distribution of expert demonstrations and have achieved strong performance across a wide range of manipulation tasks. However, a fundamental limitation of these methods lies in their slow inference, which makes it impractical to generate actions at every timestep. As a result, many approaches adopt action chunking \cite{zhao2023learning}, where a sequence of future actions is generated conditioned on the current observation and then executed without intermediate feedback. While this design enables temporally coherent behavior generation and reduces the frequency of expensive inference, the policy operates in an open-loop manner over the chunk horizon. From a control perspective, such generative imitation learning policies can be interpreted as a form of receding-horizon control, analogous to model predictive control (MPC) \cite{kouvaritakis2016model}, in which a nominal action trajectory is repeatedly replanned based on the latest observation. However, unlike classical MPC—which executes only the first control input and replans at high frequency—these methods execute multiple actions between updates due to inference latency, limiting their ability to incorporate high-frequency feedback. This limitation is particularly problematic in contact-rich manipulation tasks, where unmodeled interactions and rapid tactile feedback are critical for success. The lack of reactivity can lead to degraded performance under disturbances and model mismatch. Moreover, executing multi-step actions implicitly assumes accurate prediction of system evolution over the entire horizon, an assumption that often breaks down in contact-rich settings, resulting in brittle behaviors.

In the control literature, robustness to uncertainty has been widely studied \cite{zhou1998essentials, green2012linear, franco2006robust, shtessel2014sliding}. One representative approach is tube model predictive control (tube MPC) \cite{langson2004robust, mayne2011tube}, which decomposes control into a nominal feedforward trajectory and a stabilizing feedback controller that keeps the actual system state within a bounded tube around the nominal trajectory. This feedback mechanism enables fast local corrections in response to disturbances and modeling errors. However, classical tube MPC typically relies on linear or locally linear dynamics, requires explicit system models, and incurs substantial computational overhead, making it difficult to scale to complex, nonlinear, contact-rich manipulation tasks.

Inspired by the tube MPC paradigm, this work aims to endow diffusion-based imitation learning policies with similar robustness and reactivity by learning a local feedback flow around nominal trajectories. Instead of explicitly modeling system dynamics or deriving analytical feedback gains, we leverage the expressive capacity of generative models to directly learn an observation-conditioned feedback policy from demonstrations. To this end, we propose \emph{Tube Diffusion Policy} (TDP), a reactive visual–tactile policy learning framework that augments diffusion-based imitation learning with learned feedback control during action execution. The central idea is that, rather than committing to a fixed action sequence as \emph{action chunk}, we construct an \emph{action tube} that implicitly constrains execution within a bounded neighborhood around a nominal trajectory while allowing actions to be locally adjusted through a learned feedback flow. In particular, at the beginning of each action chunk horizon, a diffusion model generates an initial action that accounts for the full system dynamics and serves as a nominal initialization. The learned feedback flow then streams subsequent actions conditioned on fresh observations, enabling local regulation within the action tube during execution. The entire structure is learned directly from demonstrations without requiring explicit system models, making it well suited for nonlinear and contact-rich manipulation.



To reconcile expressive generative modeling with efficient feedback control within a unified policy architecture, we introduce a dual-time formulation that explicitly separates diffusion and streaming flow in policy learning. A spatial timescale governed by diffusion dynamics captures the multimodal action distribution while accounting for full contact dynamics, whereas a temporal timescale governs action evolution by streaming a learned flow defined by a velocity field. This dual-time architecture enables accurate action generation through diffusion while supporting fast observation-conditioned corrections during execution. As a result, TDP mitigates the open-loop execution inherent in action chunking approaches, while retaining the expressiveness of diffusion-based policy learning.

Overall, this framework can be viewed as a learning-based realization of tube MPC under unknown dynamics. Rather than solving a model-based optimization problem, TDP leverages a learned action velocity field to induce locally stabilizing behavior. Extensive simulation and real-world experiments on challenging visual–tactile manipulation tasks demonstrate substantial improvements over state-of-the-art imitation learning baselines in both task success and reaction under uncertainty.

The main contributions of this work are summarized as follows:
\begin{itemize}
\item We introduce \emph{action tube}, which extends the widely used \emph{action chunk} with a learned local feedback flow, enabling reactive control during execution. This formulation also relaxes the need for highly accurate action generation through denoising and thereby significantly reduces control latency.

\item We propose \emph{Tube Diffusion Policy} (TDP), a novel architecture that integrates diffusion-based action generation with learned feedback flow, enabling reactive visual–tactile policy learning for contact-rich manipulation.

\item We provide a theoretical stability analysis showing that reactive control can be formulated as learning an observation-conditioned action velocity field under locally linearized dynamics, while diffusion performs chunk-level corrections that account for full system dynamics.

\item We validate the proposed approach through extensive simulation and real-world experiments on challenging visual–tactile manipulation tasks, demonstrating consistent improvements over state-of-the-art imitation learning baselines in task success and reactivity under perturbations.
\end{itemize}

\section{Related Work}


The proposed Tube Diffusion Policy sits at the intersection of three research threads: (1) \emph{chunk-based policy learning}, which provides the temporal abstraction and generative modeling that TDP uses for action generation; (2) \emph{visual--tactile manipulation}, which motivates the need for high-frequency reactive control under contact uncertainty; and (3) \emph{learning dynamical systems}, which supplies the velocity-field formulation that TDP adopts for streaming execution. As summarized in Table~\ref{tab:dimension_comparison}, each thread individually addresses only a subset of the requirements for reactive contact-rich manipulation. We review each thread below, focusing on the specific limitation that TDP is designed to overcome.

\begin{table*}[t]
\centering
\caption{Comparison of policy learning paradigms for reactive visual-tactile manipulation.}
\label{tab:dimension_comparison}
\renewcommand{\arraystretch}{1.1}
\setlength{\tabcolsep}{4pt}
\begin{tabular}{lcccc}
\toprule
\textbf{Method} 
& \textbf{Visual-Tactile Observation} 
& \textbf{Closed-loop Execution} 
& \textbf{Contact-rich Dynamics}
& \textbf{Fast Inference}\\
\midrule

Action Chunking with Transformers (ACT) \cite{zhao2023learning} 
& \xmark
& \xmark
& \cmark 
& \cmark \\

Diffusion Policy (DP) \cite{chi2023diffusionpolicy}
& \xmark
& \xmark  
& \cmark 
& \xmark \\

Dynamical Systems \cite{khansari2010imitation, khansari2011learning, ijspeert2013dynamical, schaal2003computational}
& \xmark 
& \cmark
& \xmark 
& \cmark \\

Streaming Flow Policy (SFP) \cite{jiang2025streaming}
& \xmark 
& \xmark
& \xmark 
& \cmark \\

Reactive Diffusion Policy (RDP) \cite{xue2025reactive}
& \cmark
& \cmark 
& \cmark 
& \xmark \\

\textbf{Tube Diffusion Policy (TDP)} 
& \cmark
& \cmark
& \cmark 
& \cmark \\

\bottomrule
\end{tabular}
\end{table*}

\subsection{Chunk-based Policy Learning}
Action chunking is a concept originating in psychology that groups a sequence of actions into a single chunk and executes it as a unit \cite{lai2022action}. It has since been widely adopted to address long-horizon decision making in continuous control \cite{li2025reinforcement, zhao2023learning}.
In imitation learning, chunk-based formulations enable policies to model expert behavior at the level of short trajectory segments rather than individual actions. Early works focus on learning reusable latent skills or motion primitives \cite{Hausman2018Learning, Lynch2020Learning, Xue24RSS}. More recent approaches directly predict action chunks using sequence models. Notably, Action Chunking Transformer (ACT) \cite{zhao2023learning} demonstrates that predicting fixed-length action sequences can significantly reduce error accumulation and enable long-horizon bimanual manipulation, even on low-cost hardware platforms. However, ACT executes the predicted chunks largely in an open-loop manner, and its performance can degrade under contact uncertainty or external disturbances within each chunk.

Generative models further enhance chunk-based policies by enabling expressive multi-step action generation. They generate entire action chunks conditioned on current observations, allowing the policy to represent multimodal behaviors and capture long-horizon dependencies \cite{chi2023diffusionpolicy, zhang2024affordance, jiang2025streaming}. While effective in visuomotor manipulation, these methods typically replan only at the chunk boundary and provide limited mechanisms for correcting high-frequency deviations caused by modeling errors or contact variations during execution.


As a result, existing chunk-based and generative policy learning methods provide powerful abstractions for long-horizon planning but remain fundamentally limited by their weak intra-chunk feedback mechanisms, particularly in contact-rich settings with tactile feedback \citep{liu2025bidirectional, xue2025reactive}. Our work aims to address this issue by integrating chunk-based generative policies with structured local feedback flows, enabling systematic disturbance rejection while preserving the benefits of temporal abstraction.


\subsection{Visual-tactile Manipulation}

Visual–tactile manipulation has attracted increasing attention, particularly in contact-rich tasks that require force regulation and fine-grained interaction. Early work leveraged tactile signals for contact detection, slip estimation, and state inference, demonstrating that tactile feedback is an essential component in tasks such as grasp stabilization \cite{calandra2018more, bauza2024simple}, in-hand manipulation \cite{suresh2024neuralfeels}, and deformable object manipulation \cite{Yuan2017Shape}.

Recent advances in visuomotor policy learning \cite{chi2023diffusionpolicy, zhao2023learning, ze20243d, zhang2024affordance} have further enabled the learning of complex manipulation behaviors, motivating the integration of tactile sensing into policy learning frameworks. Most existing approaches incorporate tactile signals as additional sensory inputs by concatenating them with visual observations into a unified observation space, and train policies end-to-end using standard visuomotor learning pipelines \cite{lee2020making, hansen2022visuotactile, zhang2023visual}. While effective in many settings, such formulations often rely on computationally intensive generative models with relatively slow inference, and therefore overlook the fact that tactile sensing is inherently high-frequency and contact-driven, requiring rapid reactive responses during execution.

To better exploit tactile feedback, visual–tactile policy learning should adopt a reactive formulation. One representative approach is Reactive Diffusion Policy (RDP) \cite{xue2025reactive}, which employs a slow–fast architecture that combines diffusion-based planning with fast tactile-driven refinement, enabling improved reactivity to external disturbances in contact-rich manipulation tasks. Nevertheless, it has been primarily validated on parallel grippers rather than dexterous hands, leaving its effectiveness in high-dimensional, multi-contact manipulation settings unclear. Moreover, it still relies on slow denoising at the beginning of each action chunk, introducing latency and limiting responsiveness, which can induce periodic jerky motions similar to those observed in standard diffusion policies.

Overall, most existing visual–tactile policies either execute planned action chunks in an open-loop manner or rely on auxiliary controllers that operate in latent space, limiting their ability to support fully reactive visual–tactile control for high-DoF dexterous manipulation. In contrast, we propose TDP, which embeds reactivity directly into the diffusion policy through a streaming integration phase, enabling online action updates under an observation-conditioned vector field.

\subsection{Learning Dynamical Systems}
Learning policies and motions as dynamical systems has a long history in robotics and imitation learning. Early work represents robot behaviors as stable nonlinear dynamical systems or attractor models, enabling globally convergent motions learned from demonstrations \cite{khansari2010imitation, khansari2011learning, ijspeert2013dynamical, schaal2003computational}. These approaches impose strong structural constraints to guarantee stability, which simplifies the underlying dynamics and facilitates analysis, but limits expressiveness when dealing with highly nonlinear or contact-rich interactions.


Recent learning-based methods extend this perspective by parameterizing continuous-time dynamics with neural networks. Neural ODEs and their variants provide a flexible framework for learning nonlinear dynamical systems from data \cite{chen2018neural, rubanova2019latent, kidger2020neural}, while recent work explicitly enforces stability properties in deep dynamical systems \cite{sochopoulos2024learning}. Although more expressive, these methods are typically used as motion generators or nominal dynamics models, rather than as fully reactive controllers under complex disturbances.

In parallel, generative modeling techniques such as diffusion models, continuous normalizing flows, and flow matching reinterpret trajectory or action generation as the integration of learned vector fields \cite{song2020score, ho2020denoising, lipman2022flow, liu2022flow}. Building on this perspective, Streaming Flow Policy (SFP) \cite{jiang2025streaming} formulates policy generation as a continuous-time dynamical system evolving in action space, enabling smooth action evolution and improved coverage of multimodal behaviors. However, due to the computational cost of generative inference, these methods primarily produce open-loop or chunk-based action trajectories, with limited ability to react to execution-time disturbances.

Overall, most prior learning-based dynamical system approaches either rely on simplified or abstracted dynamics, or structure execution around chunk-based or open-loop evolution, limiting their ability to represent complex behaviors or regulate deviations arising from disturbances or contact interactions. Our approach combines the strengths of diffusion-based policies and streaming flows, where diffusion is used to capture complex nonlinear behaviors, while observation-conditioned streaming flow enables real-time, reactive control during execution.


\section{Problem Formulation}

We consider contact-rich manipulation systems with observation $\bm{o}_t = (\bm{o}_t^{o}, \bm{o}_t^{r}) \in \mathbb{O} \subseteq \mathbb{R}^{n_o}$, where $\bm{o}_t^{r}$ denotes the robot’s proprioceptive state and $\bm{o}_t^{o}$ represents the object state, which may consist of either low-dimensional object configurations or high-dimensional, multimodal sensory inputs such as visual and tactile observations. The control input $\bm{u}_t \in \mathbb{U} \subseteq \mathbb{R}^{n_u}$ specifies the commanded robot action. The system evolves according to the continuous-time dynamics
\begin{equation}
\dot{\bm{o}}(t) = f\big(\bm{o}(t), \bm{u}(t)\big),
\label{eq:dynamics}
\end{equation}
where $f(\cdot)$ captures the underlying contact-rich interaction dynamics between the robot and its environment.

To enable reactive feedback control, we adopt a tube-based control formulation. Specifically, by locally linearizing the system over a short time horizon, we can obtain the following control law with a proportional feedback correction:
\begin{equation}
\label{eq:tube_fb_simple}
\bm{u}(t) = \bar{\bm{u}}(t) + K\big(\bar{\bm{o}}(t) - \bm{o}(t)\big),
\end{equation}
where $\bar{\bm{u}}(t)$ and $\bar{\bm{o}}(t)$ denote the nominal action and observation trajectories, respectively, and the feedback gain $K \succ 0$ is held constant within the horizon. Differentiating~\eqref{eq:tube_fb_simple} with respect to time yields
\begin{equation}
\label{eq:udot_simple}
\dot{\bm{u}}(t)
= \dot{\bar{\bm{u}}}(t)
+ K\big(\dot{\bar{\bm{o}}}(t) - \dot{\bm{o}}(t)\big).
\end{equation}

We consider a neighborhood of the nominal trajectory in which the observation dynamics are differentiable and admit a first-order approximation. In particular, we apply a Taylor expansion around $(\bar{\bm{o}}(t),\bar{\bm{u}}(t))$ of system dynamics \eqref{eq:dynamics}, leading to
\begin{equation}
\label{eq:obs_lin}
    \dot{\bm{o}}(t)
    \approx
    \dot{\bar{\bm{o}}}(t)
    + A(t)\big(\bm{o}(t)-\bar{\bm{o}}(t)\big)
    + B(t)\big(\bm{u}(t)-\bar{\bm{u}}(t)\big),
\end{equation}
where
$
A(t)\triangleq \left.\frac{\partial f}{\partial \bm{o}}\right|_{(\bar{\bm{o}}(t),\bar{\bm{u}}(t))}
$
and
$
B(t)\triangleq \left.\frac{\partial f}{\partial \bm{u}}\right|_{(\bar{\bm{o}}(t),\bar{\bm{u}}(t))}.
$

Within each time interval, we adopt the widely used quasi-dynamic assumption for contact-rich systems: over the short duration of each interval, the observation error $\bm{e}(t) = \bar{\bm{o}}(t)-\bm{o}(t)$ remains bounded and varies slowly, such that its contribution can be neglected in the local approximation. This yields
\begin{equation}
\label{eq:Gamma_def}
    \dot{\bar{\bm{o}}}(t)-\dot{\bm{o}}(t)
    \approx
    -\Gamma(t)\big(\bm{u}(t)-\bar{\bm{u}}(t)\big),
    \qquad \Gamma(t)\triangleq B(t),
\end{equation}
where $\Gamma(t)$ summarizes the local input-to-observation sensitivity induced by contact interactions and the environment, and is understood to be valid locally.

Substituting~\eqref{eq:Gamma_def} into~\eqref{eq:udot_simple} yields the action-space velocity field
\begin{equation}
\label{eq:action_velocity_matrix}
    \dot{\bm{u}}(t)
    \approx
    \dot{\bar{\bm{u}}}(t)
    - K\Gamma(t)\big(\bm{u}(t)-\bar{\bm{u}}(t)\big).
\end{equation}
Equation~\eqref{eq:action_velocity_matrix} decomposes the controller into a trajectory-following term $\dot{\bar{\bm{u}}}(t)$ and a stabilizing contraction term that drives system deviations toward the nominal trajectory, thereby locally maintaining the execution error within a bounded tube. 

For simplicity, we approximate the product $K\Gamma(t)$ by an isotropic contraction term $\lambda I$. This approximation is motivated by the observation that action–observation coupling in contact-rich manipulation is often dominated by a few low-dimensional interaction modes within a local contact regime. For instance, in planar pushing with persistent contact, the interaction is typically governed by contact modes such as sticking or sliding at the contact interface \cite{Xue23ICRA, Xue25IJRR}. Within each mode, the resulting action–observation coupling is effectively low-dimensional, making an isotropic contraction a reasonable local approximation. Here, $\lambda > 0$ denotes the effective local contraction rate in action space. Substituting this approximation into~\eqref{eq:udot_simple} yields the simplified action-space dynamics
\begin{equation}
\label{eq:action_velocity_simple}
    \dot{\bm{u}}(t)
    \approx
    \dot{\bar{\bm{u}}}(t)
    - \lambda \big(\bm{u}(t) - \bar{\bm{u}}(t)\big),
    \qquad \lambda \triangleq \gamma k > 0,
\end{equation}
which describes a contracting flow around the nominal action trajectory.

This formulation is structurally consistent with~\cite{jiang2025streaming}, which frames policy learning as the imitation of action trajectories. While effective in some settings, \cite{jiang2025streaming} abstracts away robot–environment coupling and is generally insufficient, as control policies are fundamentally designed to regulate system states or observations rather than actions alone. We instead adopt a full-system perspective and identify physically grounded assumptions under which action-trajectory imitation provides a valid local approximation of closed-loop control. We show that this approximation holds only in a neighborhood of the nominal trajectory under a quasi-dynamic execution regime with bounded tracking errors. This analysis provides the theoretical motivation for the proposed approach, in which diffusion provides chunk-level corrections that account for full system dynamics, while the streaming flow enables local feedback control during execution.

\section{Methodology}

From~\eqref{eq:action_velocity_simple}, we observe that after local linearization, the resulting feedback flow depends only on the nominal action trajectory. This insight suggests that the conventional decision-making process can be conceptually decomposed into two stages: (i) periodically generating actions that account for the full nonlinear dynamics to correct accumulated drift, and (ii) applying a local feedback controller based on linearized dynamics.

Motivated by this decomposition, we propose \emph{Tube Diffusion Policy} (TDP), a framework for reactive policy learning under real-time observations. At the beginning of each action chunk, a diffusion model performs multi-step denoising to generate an initial action that accounts for the full nonlinear contact dynamics. Subsequent actions are then generated through a streaming feedback flow learned from demonstrations, which locally regulates deviations based on fresh observations. Although the diffusion stage does not explicitly generate a full nominal action trajectory, it implicitly defines one by providing the initial condition for the subsequent action evolution. The resulting action sequence can therefore be interpreted as a nominal trajectory around which closed-loop regulation is performed. This structure aligns with the action chunk paradigm in chunk-based control, but crucially augments it with step-wise closed-loop feedback during execution. We refer to this formulation as an \emph{action tube}, where diffusion provides chunk-level action initialization and the learned feedback flow enforces real-time corrections within the tube. This section presents the overall architecture for learning such reactive control policies from demonstrations. 

Given a dataset of demonstration episodes, similar to chunk-based approaches \citep{chi2023diffusionpolicy, zhao2023learning}, we decompose each episode into multiple observation and action sequences of length $H_p$, yielding a dataset $\mathcal{D} = \{(\zeta^{o}_i, \zeta^{u}_i)\}_{i=1}^N$. Our goal is to learn the conditional velocity field $v_{\theta}(\bm{u}, t_1, t_2 \mid h_{\tau})$ by minimizing~\eqref{eq:loss}, enabling both diffusion-based action generation and flow-based reactive control. Here, $h_{\tau} = \{\bm{o}_t\}_{t=\tau}^{\tau+H_o}$ denotes a finite observation history of length $H_o$. At execution time, the learned policies are deployed in a receding-horizon fashion, with an action horizon $H_a$ that may differ from the prediction horizon $H_p$ used during training. 
The complete pseudocode for training and inference is provided in Alg. \ref{alg:training} and Alg. \ref{alg:inference}.

\subsection{Overall Architecture}

\begin{figure}[t]
\centering
    \includegraphics[width=0.9\linewidth]{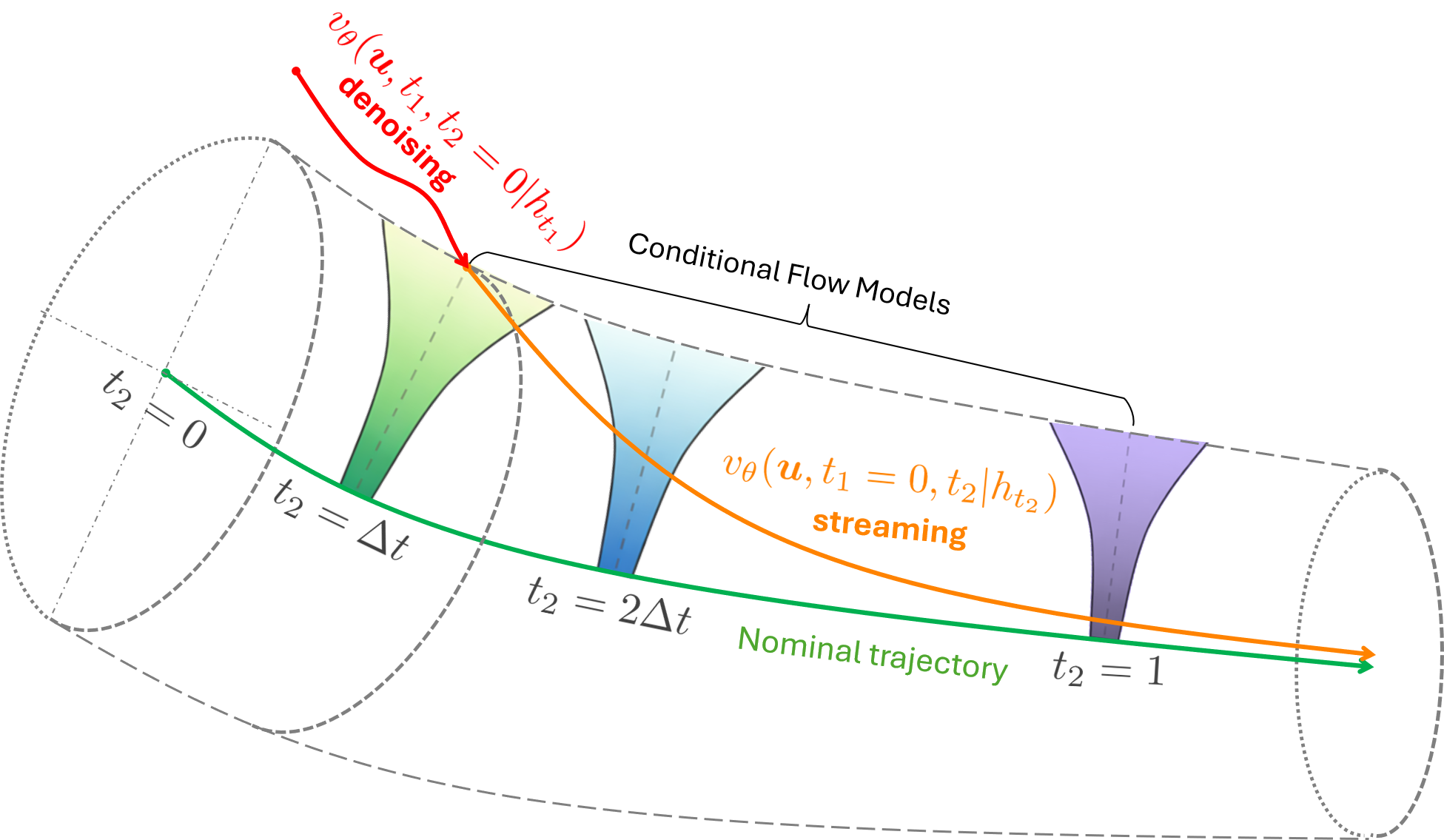}
    \caption{\textbf{Overview of Tube Diffusion Policy (TDP).} TDP adopts a dual-time formulation with a multi-step denoising stage along $t_1$ and an observation-conditioned streaming stage along $t_2$. The method operates in two phases: a \emph{denoising phase} and a \emph{streaming phase}. At the beginning of each action chunk horizon, an initial action is generated through multi-step denoising, accounting for the full nonlinear system dynamics; the resulting trajectory is shown in red. Subsequent actions are produced by streaming action flow conditioned on the observation at each timestep, yielding the orange trajectory. The green curve denotes the nominal trajectory. This process can be interpreted as executing actions within an \emph{action tube} (dashed boundary), where local deviations are continuously regulated by switching among a sequence of observation-conditioned flow models (illustrated as funnels). Here, $h_t$ denotes the observation history, and $\Delta t$ controls the streaming temporal granularity. By combining diffusion-based chunk-level initialization with tube-based reactive feedback, TDP enables robust, high-frequency closed-loop control for contact-rich manipulation.}
    
    \label{fig:overview}
\end{figure}

\paragraph{Dual-Time Formulation}
TDP introduces two time variables with distinct semantics:
\begin{itemize}
\item $t_1 \in [0, T_{\text{diff}}]$: \textbf{diffusion time}, governing the diffusion denoising process while the physical system remains stationary;
\item $t_2 \in [0, 1]$: \textbf{trajectory time}, governing the streaming control process that evolves in real time with the physical system.
\end{itemize}
The policy is conditioned on both variables, allowing a single network to operate in two modes: diffusion-based action generation and flow-based streaming.

\paragraph{Training Objective}
The overall training objective combines losses from both modes:
\begin{equation}
\mathcal{L} = \mathcal{L}_{\text{diff}} + \mathcal{L}_{\text{stream}},
\label{eq:loss}
\end{equation}
where $\mathcal{L}_{\text{diff}}$ is the diffusion noise-prediction loss and $\mathcal{L}_{\text{stream}}$ supervises the streaming velocity field. We use equal weighting for the two objectives, as they correspond to complementary roles—global action generation and local feedback correction—and operate on comparable scales in practice.

\subsection{Diffusion Phase: Diffusion-based Action Generation}

The diffusion phase follows the standard Denoising Diffusion Probabilistic Models (DDPM) formulation with $\epsilon$-prediction~\cite{ho2020denoising}.
Given a clean action $\bm{u}_0=\zeta^u[H_o -1:H_o]$ that is aligned with the observation history $h_{t_1} = \zeta^o[:H_o]$,
the forward noising process is defined as
\begin{equation}
\bm{u}_{t_1}
= \sqrt{\bar{\alpha}_{t_1}} \bm{u}_0
+ \sqrt{1 - \bar{\alpha}_{t_1}} \boldsymbol{\epsilon},
\qquad \boldsymbol{\epsilon} \sim \mathcal{N}(\mathbf{0}, \mathbf{I}),
\label{eq:diffused_action}
\end{equation}
where $\bar{\alpha}_{t_1}$ is specified by the noise schedule in~\cite{ho2020denoising}.
The network $v_{\theta}(\bm{u}, t_1, t_2 \mid h_{t_1})$ is trained to predict the injected noise
conditioned on the aligned observation history $h_{t_1}$ by minimizing
\begin{equation}
\mathcal{L}_{\mathrm{diff}}
=
\mathbb{E}_{\bm{u}_0,\, h_{t_1},\, \boldsymbol{\epsilon},\, t_1}
\Big[
\big\|
\boldsymbol{\epsilon}
-
v_{\theta}(\bm{u}_{t_1}, t_1, t_2 = 0 \mid h_{t_1})
\big\|^2
\Big],
\label{eq:diffusion_loss}
\end{equation}
where the expectation is taken over aligned observation--action pairs
$(h_{t_1}, \bm{u}_0)$ sampled from the demonstration dataset $\mathcal{D}$, and
$t_1$ is sampled uniformly from $[0,T_{\text{diff}}]$, with $[0,T_{\text{diff}}]$ specifying the length of the diffusion process. 


\begin{figure*}[t]
\centering
\begin{minipage}[t]{0.48\textwidth}
\begin{algorithm}[H]
\caption{Training}
\label{alg:training}
\begin{algorithmic}[1]
\REQUIRE Training set $\mathcal{D} =\{(\zeta^{o}_i, \zeta^{u}_i)\}_{i=1}^N$, $H_o$, $H_p$, $T_{\text{diff}}$
\STATE $\zeta^{o}$ and $\zeta^{u}$ have time horizon $T_{\text{pred}}$ rescaled to $[0, 1]$
\WHILE{not converged}
    \STATE $(\zeta^o,\zeta^u) \sim \mathcal{D}$
    \STATE \textcolor{commentblue}{\textit{// ===== Diffusion phase ($t_1$) =====}}
    \STATE $t_1 \sim \text{Uniform}\{0, \ldots, T_{\text{diff}}\}$ \hfill \textcolor{commentblue}{\textit{// diffusion time}}
    \STATE $h_{t_1} \gets \zeta^{o}[:H_o]$
    \STATE $\bm{u}_0 \gets \zeta^{u}[H_o-1:H_o]$
    \STATE $\boldsymbol{\epsilon} \sim \mathcal{N}(\mathbf{0}, \mathbf{I})$
    \STATE $\bm{u}_{t_1} = \sqrt{\bar{\alpha}_{t_1}} \bm{u}_0 + \sqrt{1 - \bar{\alpha}_{t_1}} \boldsymbol{\epsilon}$ \hfill \textcolor{commentblue}{\textit{// Eq. \eqref{eq:diffused_action}}}
    \STATE $\mathcal{L}_{\text{diff}} \gets \|\boldsymbol{\epsilon} - v_{\theta}(\bm{u}_{t_1}, t_1, t_2 = 0 \mid h_{t_1})\|^2$  \hfill \textcolor{commentblue}{\textit{// Eq. \eqref{eq:diffusion_loss}}}
    \STATE \textcolor{commentblue}{\textit{// ===== Streaming phase ($t_2$) =====}}
    \STATE $t_2 \sim \text{Uniform}(0, 1)$ \hfill \textcolor{commentblue}{\textit{// trajectory time}}
    \STATE $\bm{u} \sim p_{\zeta^{u}}(\bm{u} \mid t_2)$
    \STATE $l \gets \lfloor t_2 H_p \rfloor$
    \STATE $h_{t_2} \gets \zeta^{o}[l:l + H_o]$ \hfill \textcolor{commentblue}{\textit{// Eq. \eqref{eq:hist_obs}}}
    \STATE $v_{\zeta}(\bm{u}, t_2) \gets \textit{Eq. \eqref{eq:stream_flow}}$
    \STATE $\mathcal{L}_{\text{stream}} \gets \|v_{\zeta}(\bm{u}, t_2) - v_\theta(\bm{u}, t_1 =0, t_2| h_{t_2})\|^2$
    \STATE $\theta \gets \theta - \gamma \nabla_\theta (\mathcal{L}_{\text{diff}} + \mathcal{L}_{\text{stream}})$
\ENDWHILE
\RETURN $v_\theta$
\end{algorithmic}
\end{algorithm}
\end{minipage}
\hfill
\begin{minipage}[t]{0.48\textwidth}
\begin{algorithm}[H]
\caption{Inference}
\label{alg:inference}
\begin{algorithmic}[1]
\REQUIRE $v_\theta(\bm{u}, t_1, t_2 \,|\, h)$, $H_o$, $H_p$, $H_a$, $T_{\text{ddim}}$, $\Delta t$
\STATE $h \gets \{ \bm{o}_{i} \}_{i=1}^{H_o}$ \hfill \textcolor{commentblue}{\textit{// observation history}}
\WHILE{True}
    \STATE \textcolor{commentblue}{\textit{// ===== Phase 1: Denoising ($t_1$)=====}}
    \STATE  $\bm{u} \sim \mathcal{N}(\mathbf{0}, \mathbf{I})$
    \FOR{$t_1 = T_{\text{ddim}}, \ldots, 1$}
        \STATE $\hat{\boldsymbol{\epsilon}} \gets v_\theta(\bm{u}, t_1, t_2=0 \,|\, h)$
        \STATE $\bm{u} \gets \text{DDIMStep}(\bm{u}, \hat{\boldsymbol{\epsilon}}, t_1)$ 
    \ENDFOR
    \STATE \textcolor{commentblue}{\textit{// ===== Phase 2: Streaming ($t_2$) =====}}
    \STATE $t_2 \gets \Delta t$
    \WHILE{$t_2 \leq H_a / H_p$}
        \STATE $o \gets \text{Execute}(\bm{u})$ \hfill \textcolor{commentblue}{\textit{// stream action}}
        \STATE $h \gets h \cup \{o\}$ \hfill \textcolor{commentblue}{\textit{// fresh observation}}
        \STATE $\bm{u} \gets \bm{u} + v_\theta(\bm{u}, t_1 = 0, t_2 \,|\, h) \cdot \Delta t$ \hfill \textcolor{commentblue}{\textit{// integration}}
        \STATE $t_2 \gets t_2 + \Delta t$
    \ENDWHILE
\ENDWHILE
\end{algorithmic}
\end{algorithm}
\end{minipage}
\end{figure*}

\subsection{Streaming Phase: Flow-based Action Streaming}

By locally linearizing the system dynamics, the tube-based feedback controller can be projected from the observation–action space into the action space. Under this formulation, the streaming phase learns a reactive controller by directly modeling action evolution as a velocity field over the trajectory time, enabling high-frequency closed-loop execution. This design is inspired by~\cite{jiang2025streaming}.

Following~\eqref{eq:action_velocity_simple}, we define the target velocity field along the action trajectory as
\begin{equation} 
v_{\zeta}(\bm{u}, t_2) = \frac{d \zeta^{u}}{d t_2}  -\lambda \big( \bm{u} - \zeta^{u}(t_2) \big) , 
\label{eq:stream_flow} 
\end{equation}
where $\zeta^{u}(t_2)$ denotes the continuous nominal action trajectory parameterized by the trajectory time $t_2 \in [0,1]$. The first term provides feedforward progression along it, while the second induces feedback toward the nominal trajectory.

To enable reactive control within the action tube, we extract step-wise observation history aligned with the trajectory time via
\begin{equation}
h_{t_2} = \zeta^{o}[l:l+H_o],
\label{eq:hist_obs}
\end{equation}
where $l = \lfloor t_2 H_p \rfloor$ is the index corresponding to $t_2$ on the discretized observation trajectory, with $\lfloor \cdot \rfloor$ denoting the floor (round-down) operation. $H_p$ denotes the prediction horizon, and $H_o$ is the length of the observation history window.

The network $v_{\theta}(\bm{u}, t_1, t_2 \mid h_{t_2})$ is trained to predict the streaming velocity field conditioned on the aligned observation history $h_{t_2}$ by minimizing 
\begin{equation} 
\mathcal{L}_{\mathrm{stream}} = \mathbb{E}_{\bm{u},\, h,\, t_2} 
\Big[ \big\| v_{\theta}(\bm{u}, t_1 = 0, t_2 \mid h_{t_2}) - v_{\zeta}(\bm{u}, t_2) \big\|^2 \Big], 
\end{equation} 
where $t_2 \sim \mathcal{U}([0,1])$, and $\bm{u}$ is sampled from the stabilizing conditional distribution
\begin{equation} 
\bm{u} \sim p_{\zeta}(\bm{u} \mid t_2) = \mathcal{N}\!\left( \zeta^{u}(t_2), \; \sigma_0^2 e^{-2 k t_2} \mathbf{I} \right), 
\end{equation}
which is a Gaussian centered at the action trajectory $\zeta^{u}(t_2)$ with a small variance $\sigma_0^2$, as introduced in \cite{jiang2025streaming}.

By extracting step-wise observation histories, we learn a family of observation-conditioned flow models of the form $v_{\theta}(\bm{u}, t_1=0, t_2 \mid h_{t_2})$, each conditioned on a distinct observation history $h_{t_2}$. During execution, the action sequence within the chunk horizon is not governed by a single conditional flow model; instead, actions are generated by sequentially switching among multiple observation-conditioned flows as new observation becomes available and provide updated conditioning.

At each step, the action is updated by integrating the currently active streaming velocity field,
\begin{equation}
\bm{u} \leftarrow \bm{u} + v_{\theta}(\bm{u}, t_1=0, t_2 \mid h_{t_2})\, \Delta t_2,
\end{equation}
where $\Delta t_2$ denotes the discretization step size of the trajectory time $t_2$. 
This step-wise integration incorporates fresh observations at every step, enabling closed-loop and reactive control.

\subsection{Network Architecture and Encoding}

We employ a conditional 1D U-Net as the shared backbone for both diffusion-based action generation and streaming-based reactive flow. The network predicts action evolution conditioned on observations, outputting either diffusion noise or action velocity depending on the time conditioning, which enables parameter sharing across the diffusion and streaming phases.

To support the dual-time formulation, the network uses separate sinusoidal embeddings for the diffusion time $t_1$ and the streaming time $t_2$. The two embeddings are scaled differently to reflect their distinct semantics: diffusion time follows the standard DDPM time scale, while trajectory time is amplified to provide finer resolution over the normalized execution horizon $[0,1]$. This separation prevents interference between diffusion and streaming behaviors while allowing a unified architecture.

Observation features and time embeddings are injected into the network via Feature-wise Linear Modulation (FiLM)~\cite{perez2018film}. This conditioning mechanism enables the policy to adapt its action predictions to both the current observation context and the operating mode (diffusion or streaming) in a smooth and modular manner.

Visual observations are encoded using ResNet-18 backbones \cite{he2016deep}. In simulation, tactile inputs are processed using a lightweight multilayer perceptron, as the tactile sensor provides array-based measurements. In real-world experiments, tactile observations from Digit360 sensors are image-based and are therefore encoded using multiple ResNet-18 backbones. The resulting features are concatenated with proprioceptive signals to form a global observation representation, which conditions the action policy through FiLM modulation. Detailed network architecture and hyperparameters are provided in Appendix~\ref{app:architecture}.

Actions consist of end-effector position, orientation, and hand joint commands. For orientation, we adopt the continuous 6D rotation representation~\cite{zhou2019continuity}, which avoids discontinuities associated with quaternion parameterizations and is better suited for regression-based learning.

\subsection{Stability Analysis}

TDP decomposes control into two complementary mechanisms: diffusion correction, which periodically contracts the accumulated drift, and streaming imitation, which provides high-frequency reactive control but may introduce bounded tracking error. Together they form a hybrid control system that admits a practical stability guarantee.

Consider the discrete-time system
\begin{equation}
\bm{o}_{t+1} = g(\bm{o}_t,\bm{u}_t) + \bm{w}_t ,
\end{equation}
where \(g(\cdot)\) denotes the discrete-time dynamics corresponding to \eqref{eq:dynamics}, and \(\bm{w}_t\) is a bounded disturbance.

Let \((\bm{o}_t^*,\bm{u}_t^*)\) be a demonstrated reference trajectory satisfying
\begin{equation}
\bm{o}_{t+1}^* = g(\bm{o}_t^*,\bm{u}_t^*),
\end{equation}
and define the tracking error \(\bm{e}_t = \bm{o}_t - \bm{o}_t^*\) and the value function \(V(\bm{e}_t) = \|\bm{e}_t\|\).

To facilitate the stability analysis, we introduce the following conditions. 

\textbf{Assumption 1 (Lipschitz dynamics).}
The discrete-time dynamics satisfy
\begin{equation}
\|g(\bm{o}_1,\bm{u}_1)-g(\bm{o}_2,\bm{u}_2)\|
\le L_x\|\bm{o}_1-\bm{o}_2\| + L_u\|\bm{u}_1-\bm{u}_2\|.
\end{equation}

The Lipschitz condition bounds the sensitivity of the system dynamics to state and action deviations, enabling tractable propagation of tracking errors.

\textbf{Assumption 2 (Streaming imitation accuracy).}
The streaming flow policy is trained to imitate the reference action trajectory. As a result, the action error remains bounded:
\begin{equation}
\|\bm{u}_t-\bm{u}_t^*\| \le \varepsilon_a .
\end{equation}

This condition reflects the training objective of the streaming phase in TDP and is expected to hold when the policy accurately reproduces the demonstrated actions.

\textbf{Assumption 3 (Bounded disturbance).}
The external disturbance is bounded:
\begin{equation}
\|\bm{w}_t\| \le \bar w.
\label{eq:disturbance_bound}
\end{equation}

\textbf{Assumption 4 (Diffusion correction property).}
At correction steps \(t_k\), the diffusion module produces a corrective action that reduces the accumulated tracking error:
\begin{equation}
V(\bm{e}_{t_k+1})
\le
\lambda V(\bm{e}_{t_k}) + \varepsilon_d,
\qquad 0<\lambda<1 .
\end{equation}

This condition corresponds to the training objective of the diffusion phase in TDP, which aims to contract the tracking error at correction steps by accounting for the full system dynamics.

\textbf{Lemma 1 (Streaming-phase bound).}
During streaming execution, the value function satisfies
\begin{equation}
V(\bm{e}_{t+1})
\le
L_x V(\bm{e}_t) + L_u\varepsilon_a + \bar w .
\end{equation}

\textit{Proof.}
From the system dynamics,
\[
\bm{e}_{t+1}=g(\bm{o}_t,\bm{u}_t)+\bm{w}_t-g(\bm{o}_t^*,\bm{u}_t^*).
\]
Applying Lipschitz continuity and Assumption~2 yields
\[
\|\bm{e}_{t+1}\|
\le
L_x\|\bm{e}_t\| + L_u\varepsilon_a + \bar w .
\]
Substituting the definition of \(V\) proves the result.
\hfill $\square$

The lemma shows that the streaming policy may introduce bounded tracking drift due to imitation error and disturbances, but the error growth remains bounded at each control step.

\textbf{Theorem 1 (Practical stability of TDP).}
Suppose diffusion correction occurs every \(H_a\) steps and \(L_x<1\).
Then the closed-loop system under TDP satisfies the following properties:

(1) Uniform ultimate boundedness.
The tracking error sequence is uniformly ultimately bounded.

(2) Bounded error within each diffusion cycle.
During each correction cycle \(t\in\{t_k,\dots,t_k+H_a-1\}\), the tracking error remains bounded as a function of the post-correction error.

\textit{Proof sketch.}
Lemma~1 shows that during streaming execution the value function grows at most linearly due to imitation error and disturbances. 
At correction instants, diffusion reduces the accumulated tracking error according to Assumption~4. 
Combining bounded growth with periodic contraction yields a stable affine recursion on the post-correction error, which implies uniform ultimate boundedness of the tracking error. 
Moreover, the streaming bound ensures that the error remains bounded throughout each diffusion cycle.
\hfill $\square$

In the ideal case where the imitation error, disturbance, and diffusion residual vanish
(\(\varepsilon_a = 0\), \(\bar w = 0\), and \(\varepsilon_d = 0\)),
the equilibrium \(\bm{e}=0\) is asymptotically stable and the tracking error satisfies $\lim_{t\to\infty} \bm{e}_t = 0$.
The full derivation of the stability bound is provided in Appendix.

\section{Experiments}
\label{sec:experiments}
We first use a toy 1-D point-mass system to illustrate the effectiveness of \emph{action tube} under external disturbances. We then evaluate TDP on the widely used Push-T task and three additional visual–tactile dexterous manipulation tasks in simulation, followed by two challenging real-world experiments that require reactive control under contact uncertainty and external disturbances.

\subsection{1-D Point-mass System}

\begin{figure*}[hbpt]
\centering
    \includegraphics[width=\linewidth]{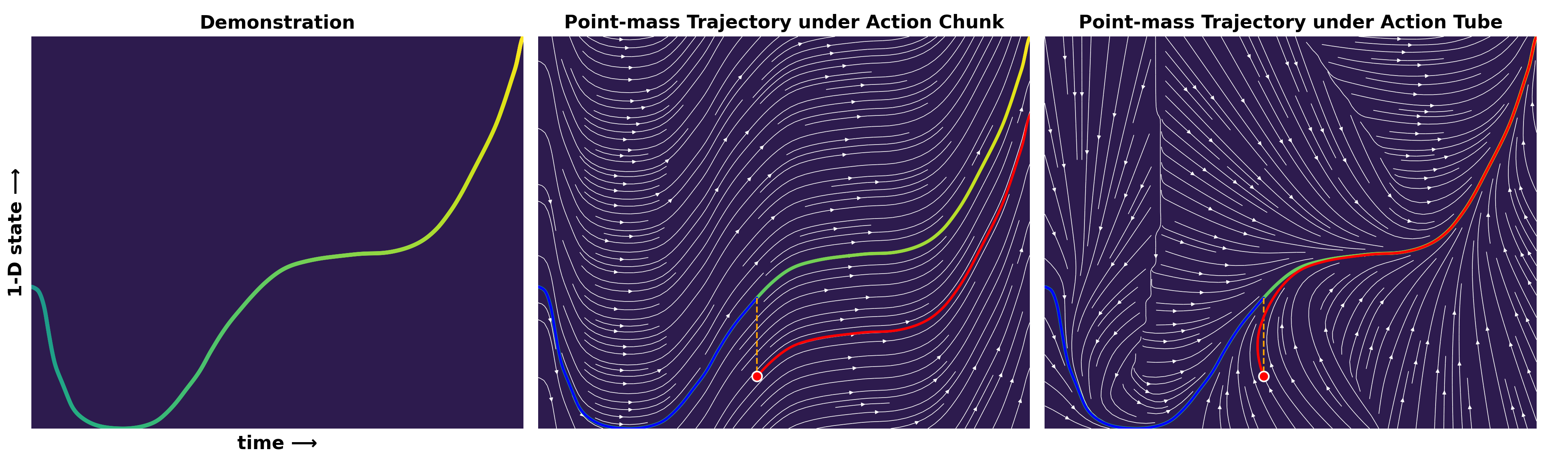}
    \caption{\textbf{1-D Point-Mass Example: Action-Tube Robustness to External Disturbances}. 
    We consider a 1-D point-mass system with a single demonstration to contrast action chunk and action tube. The x-axis denotes execution time and the y-axis denotes system state. 
    \textbf{Left:} Nominal trajectory without disturbance, with color indicating progression along the trajectory from dark green (start) to light green (end).
    \textbf{Middle:} Action chunk execution. A disturbance is applied at the midpoint (orange dashed line), after which the system continues executing the precomputed action sequence (i.e., the 1-D velocity) without feedback, resulting in deviation from the target. 
    \textbf{Right:} Action tube execution under the same disturbance. A local feedback controller drives the system back toward the nominal trajectory and ultimately reaches the target. Blue curves denote trajectories before disturbance, and red curves denote trajectories after disturbance.}
    \label{fig:1D_toy}
\end{figure*}

We consider a 1-D point-mass system with single-integrator dynamics $x_{t+1} = x_t + u_t \Delta t$. The goal is to use imitation learning to reproduce a demonstrated trajectory, as illustrated in Fig. \ref{fig:1D_toy}. The horizontal axis denotes time, while the vertical axis shows the evolution of the state $x_t$. To highlight the difference between \emph{action chunk} and \emph{action tube}, we assume that the entire trajectory is executed within a single action-chunk horizon. The white curves depict the movement flow of the point mass under different initializations in both cases. Without loss of generality, we consider the example in which the system is initialized at the same state as the demonstration, and an external disturbance is injected at the midpoint of the trajectory, as indicated by the orange dashed line. This disturbance can be interpreted as either human perturbation or physical uncertainty.

Chunk-based approaches generate a fixed sequence of actions at the beginning of execution. As a result, after the disturbance occurs, the controller continues to apply the precomputed action sequence, ultimately driving the system to an incorrect final state. In contrast, the action-tube formulation employs a local feedback controller around the nominal trajectory, which actively steers the system back toward the desired path after the disturbance.

This toy example illustrates the importance of \emph{action tube} for achieving reactive and robust control in the presence of external disturbances or contact uncertainty.

\subsection{Push-T}

We further benchmark TDP on Push-T task against state-of-the-art imitation learning methods, including Diffusion Policy (DP)~\cite{chi2023diffusionpolicy} with 100 DDPM steps, DP with 10 Denoising Diffusion Implicit Models (DDIM) \cite{song2021denoising} steps, Flow Matching Policy \cite{zhang2024affordance}, and Streaming Flow Policy (SFP) \cite{jiang2025streaming}. We also have one ablation study that removes the diffusion phase of TDP, namely TDP (streaming only). Following \cite{jiang2025streaming}, we use the current robot configuration as a warm-start initialization for SFP, and also apply the same strategy to the TDP (streaming only) ablation, as it does not include an initial denoising stage. Push-T is a widely-used task to benchmark imitation learning approaches, and it is particularly suitable to be used in this work since it has nonlinear dynamics and physical uncertainty. For example, the friction coefficients are hard to be identified and varied with different objects. Following \citep{chi2023diffusionpolicy}, we report the average score of the 5 best checkpoints and also the maximum score across all checkpoints. Moreover, there is another metric called \emph{steps to success}, which describes the task completion time for the successful trials. This further evaluates the effectiveness of each action across all cases that ultimately lead to success. We report the average steps of the 5 best checkpoints and the minimum steps of the best checkpoint. Table \ref{tab:pushT_results} demonstrates the results for both state-based and image-based Push-T tasks, where state-based Push-T task uses low-dim state as the input, including 2-D position of the cylindrical pusher and 3-D pose of the T-shape block on planar space. Image-based push-T task use RGB image as the input. The actions are 2-D setpoints tracked by a PD controller. This task contains 200 demonstrations and 50 evaluation episodes. 

As reported in Table~\ref{tab:pushT_results}, TDP achieves the highest performance on both tasks while requiring the fewest steps, demonstrating the effectiveness of real-time reactions at each step given the current observation for task accomplishment. Moreover, the ablation results show that removing the diffusion phase degrades performance, highlighting the limitations of relying solely on local linearization without a diffusion-based correction that accounts for the full system dynamics. This observation is consistent with the results of SFP, which can also be interpreted as relying primarily on locally linearized dynamics without global correction. Compared to diffusion-based baselines such as DDPM and DDIM, which achieve competitive success rates but require more steps, TDP attains higher success rates with fewer steps, highlighting the effectiveness of step-wise reactions under contact uncertainty. In contrast, flow matching shows inferior performance, reflecting its relatively limited capacity to model complex contact-rich behaviors.

\begin{figure}
    \centering
    \includegraphics[width=\linewidth]{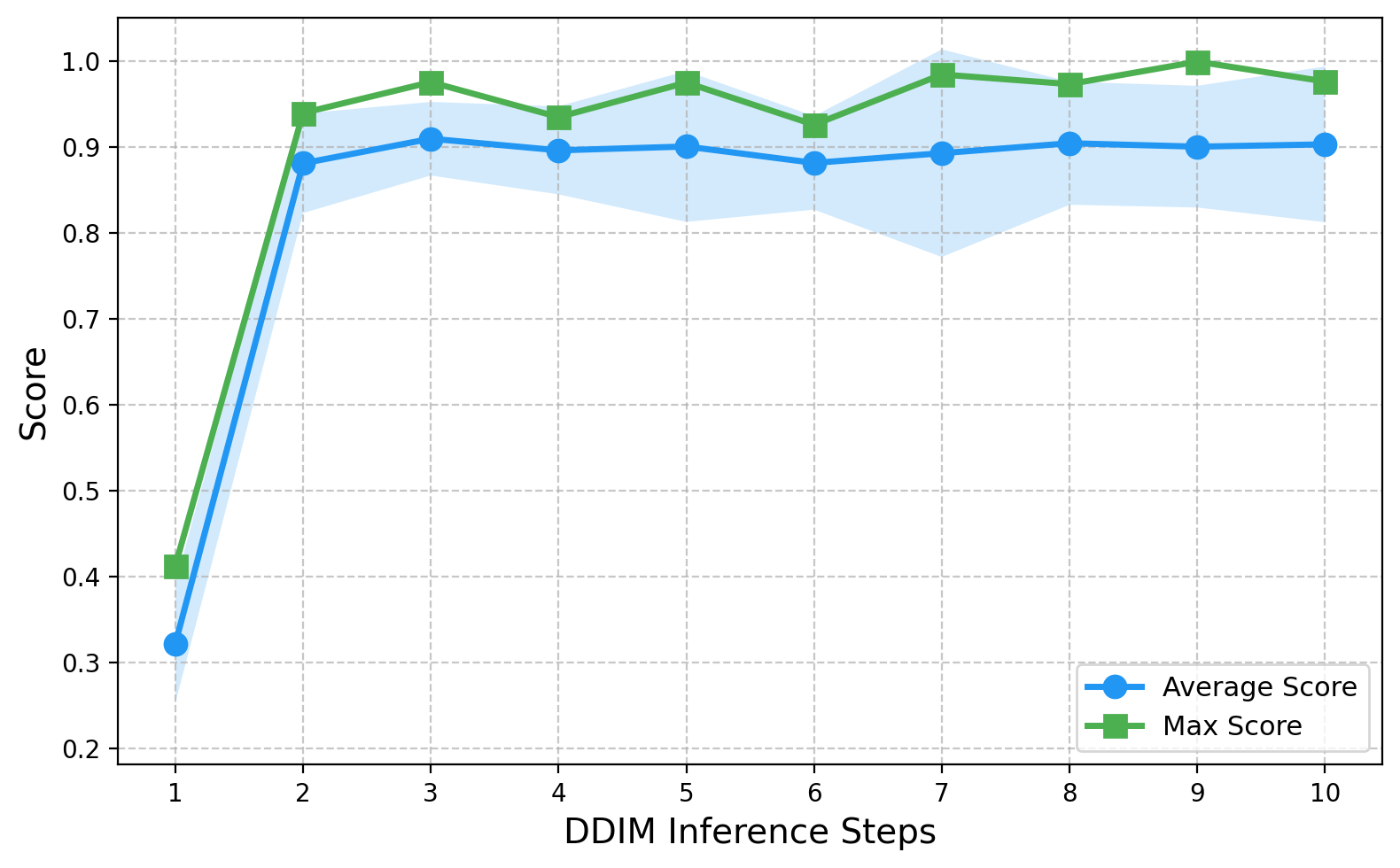}
    \caption{\textbf{Ablation study on the number of DDIM denoising steps within TDP for the virtual Push-T task.} We vary the number of DDIM inference steps used in the denoising phase of TDP. The results show that TDP maintains competitive performance with as few as two denoising steps, highlighting its suitability for high-frequency control.}
    \label{fig:push_ddim_ablation}
\end{figure}

We further conduct an ablation study on the number of DDIM inference steps used in the denoising phase for state-based Push-T task, as shown in Fig.~\ref{fig:push_ddim_ablation}. We conduct five trials, each consisting of 20 episodes, and report both the average and maximum scores. The results indicate that competitive performance can already be achieved with as few as two DDIM steps. This observation suggests that our method does not require highly accurate action generation during the denoising stage. Instead, the diffusion model only needs to produce an initial action that initializes execution while accounting for the full system dynamics, after which the streaming policy regulates deviations through closed-loop feedback. This property highlights a key advantage of the proposed tube-based formulation over conventional chunk-based methods that rely on precise action chunks. Moreover, reducing the number of DDIM steps substantially lowers inference latency, demonstrating the potential of TDP for real-time, high-frequency control.

\begin{figure}[t]
\centering
    \includegraphics[width=\linewidth]{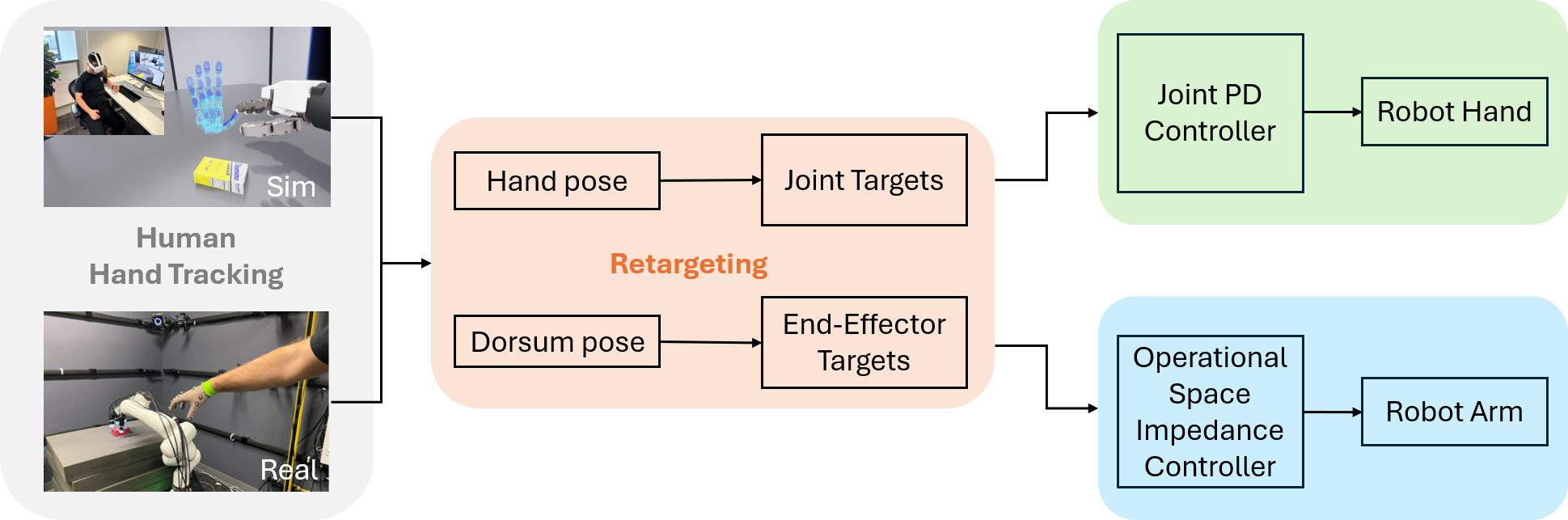}
    \caption{\textbf{Teleoperation system}. Human demonstrations are collected through teleoperation in both simulation and the real world. In simulation, a Meta Quest 3 headset is used for VR-based hand tracking, while in the physical setup an OptiTrack motion-capture system provides mocap-based hand tracking. The captured hand pose and dorsum pose are then processed by the same retargeting module and controllers to actuate the dexterous hand and robot arm.}
    \label{fig:teleopt_pipe}
\end{figure}

\begin{table*}[htbp]
\centering
\renewcommand{\arraystretch}{1.35}
\setlength{\tabcolsep}{10pt}

\definecolor{baseline}{RGB}{246,224,202}   
\definecolor{ablation}{RGB}{201,216,242}   
\definecolor{ours}{RGB}{215,242,214}       

\begin{tabular}{lcccc}
\toprule
\multirow{3}{*}{%
  \begin{minipage}[c]{0.14\linewidth}\centering
  \includegraphics[width=0.9\linewidth]{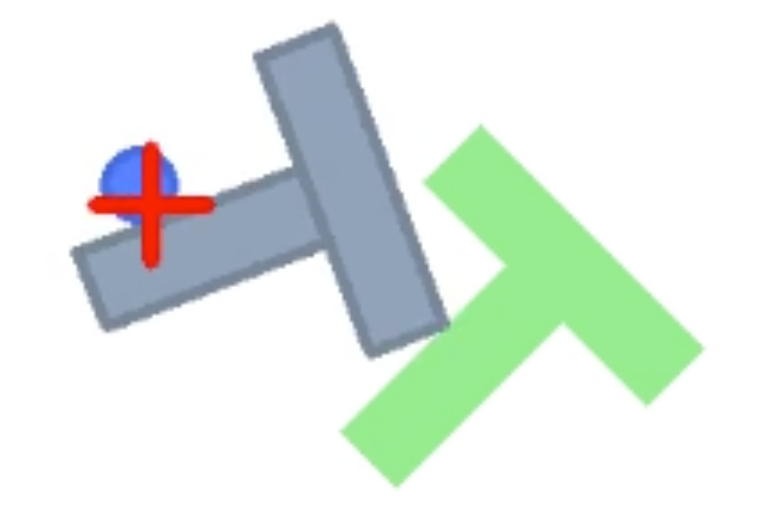}
  \end{minipage}
}
&
\multicolumn{2}{c}{\textbf{Push-T with state input}}
&
\multicolumn{2}{c}{\textbf{Push-T with image input}}
\\
\cmidrule(lr){2-3}\cmidrule(lr){4-5}
&
\textbf{Avg/Max scores} &
\textbf{Avg/Min steps to success} &
\textbf{Avg/Max scores} &
\textbf{Avg/Min steps to success}
\\
&
$\uparrow$ & $\downarrow$ & $\uparrow$ & $\downarrow$
\\
\midrule

\rowcolor{baseline}
DP (DDPM) \cite{chi2023diffusionpolicy} &
93.2\%/93.8\% &
125.9/128.5 &
76.0\%/77.5\% &
114.7/107.9
\\

\rowcolor{baseline}
DP (DDIM) \cite{chi2023diffusionpolicy} &
91.8\%/93.9\% &
124.6/117.1 &
68.5\%/69.0\% &
107.3/104.8
\\

\rowcolor{baseline}
Flow Matching \cite{zhang2024affordance} &
91.8\%/93.3\% &
122.8/115.9 &
65.3\%/68.3\% &
108.9/102.5
\\

\rowcolor{baseline}
SFP \cite{jiang2025streaming}&
88.6\%/89.7\% &
124.4/116.1 &
73.3\%/74.8\% &
112/108.6
\\

\midrule
\rowcolor{ablation}
TDP (streaming only) &
84.3\%/85.0\% &
109.6/108.1 &
77.1\%/77.7\% &
117.5/96.3
\\

\midrule
\rowcolor{ours}
\textbf{TDP} &
\textbf{96.1\%/96.9\%} &
\textbf{106.2/105.1} &
\textbf{82.0\%/85.6\%} &
\textbf{103.5/91.3}
\\

\bottomrule
\end{tabular}

\caption{\textbf{Virtual push-T results with state and image inputs.} Our method (green) is compared against baselines (orange) and ablation variants (blue).}
\label{tab:pushT_results}
\end{table*}

\begin{table*}[htbp]
\centering
\renewcommand{\arraystretch}{1.35}
\setlength{\tabcolsep}{10pt}

\definecolor{baseline}{RGB}{246,224,202}   
\definecolor{ablation}{RGB}{201,216,242}   
\definecolor{ours}{RGB}{215,242,214}       

\begin{tabular}{lccccc}
\toprule
\multirow{3}{*}{%
  \begin{minipage}[c]{0.18\linewidth}\centering
  \includegraphics[width=\linewidth]{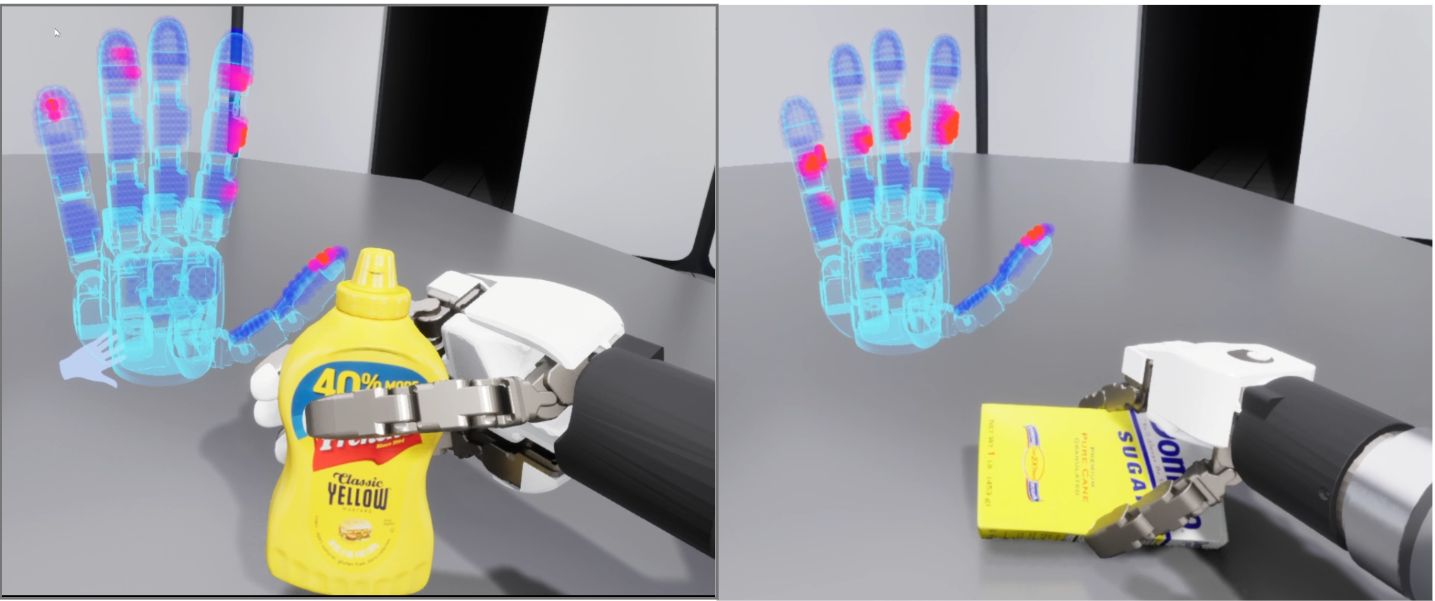}
  \end{minipage}
}
&
\multicolumn{3}{c}{\textbf{Stable Grasping}}
&
\multicolumn{2}{c}{\textbf{On-table Reorientation}}
\\
\cmidrule(lr){2-4}\cmidrule(lr){5-6}
&
\textbf{Success rate (Grasp)} &
\textbf{Success rate (Stable grasp)} &
\textbf{Steps to success} &
\textbf{Success rate} &
\textbf{Steps to success}
\\
&
$\uparrow$ & $\uparrow$ & $\downarrow$ & $\uparrow$ & $\downarrow$
\\
\midrule

\rowcolor{baseline}

\rowcolor{baseline}
DP (DDIM) \cite{chi2023diffusionpolicy}&
88\% &
60\% &
$28.0 \pm 3.70$ &
82\% &
$41.7 \pm 8.51$
\\

\rowcolor{baseline}
SFP \cite{jiang2025streaming}&
78\% &
72\% &
$29.6 \pm 4.75$ &
80\% &
$43.1 \pm 12.9$
\\
\midrule
\rowcolor{ablation}
TDP (streaming only) &
72\% &
44\% &
\textbf{25.5} $\pm$ \textbf{5.1} &
76\% &
$60.6 \pm 25.3$
\\

\midrule
\rowcolor{ours}
\textbf{TDP} &
\textbf{96\%} &
\textbf{88\%} &
$27.8 \pm 3.06$ &
\textbf{90\%} &
\textbf{40.0} $\pm$ \textbf{15.3}
\\

\bottomrule
\end{tabular}

\caption{\textbf{Comparison on virtual stable grasping and on-table reorientation.} Our method (green) is compared against baselines (orange) and ablation variants (blue).}
\label{tab:grasp_reorient_results}
\end{table*}

\subsection{Visual–Tactile Manipulation in a Virtual Environment}

\begin{figure*}[htbp]
    \centering
    \begin{subfigure}[b]{0.48\textwidth}
        \centering
        \includegraphics[width=\textwidth]{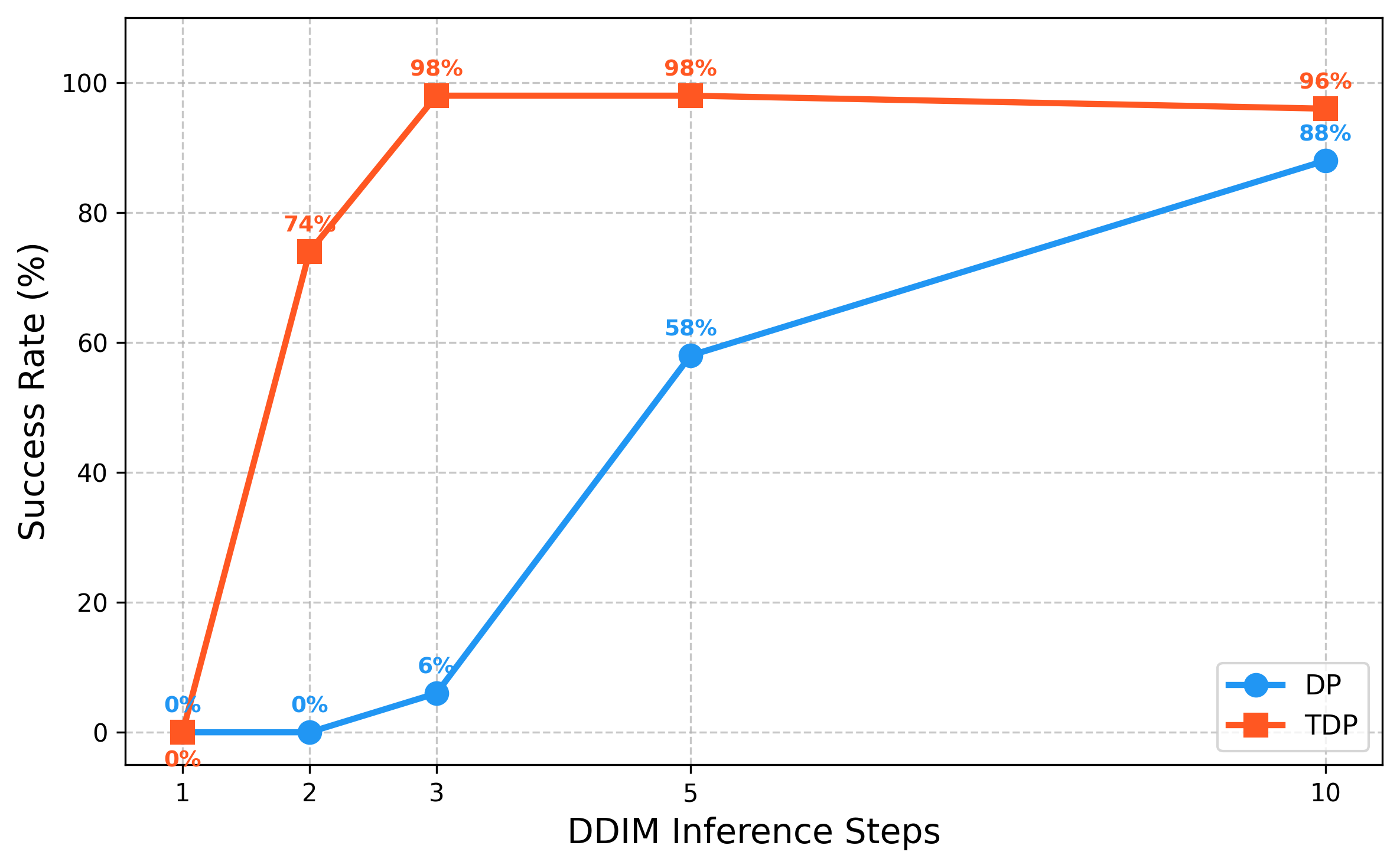}
        \caption{Grasping}
        \label{fig:ddim_ablation_grasping}
    \end{subfigure}
    \hfill
    \begin{subfigure}[b]{0.48\textwidth}
        \centering
        \includegraphics[width=\textwidth]{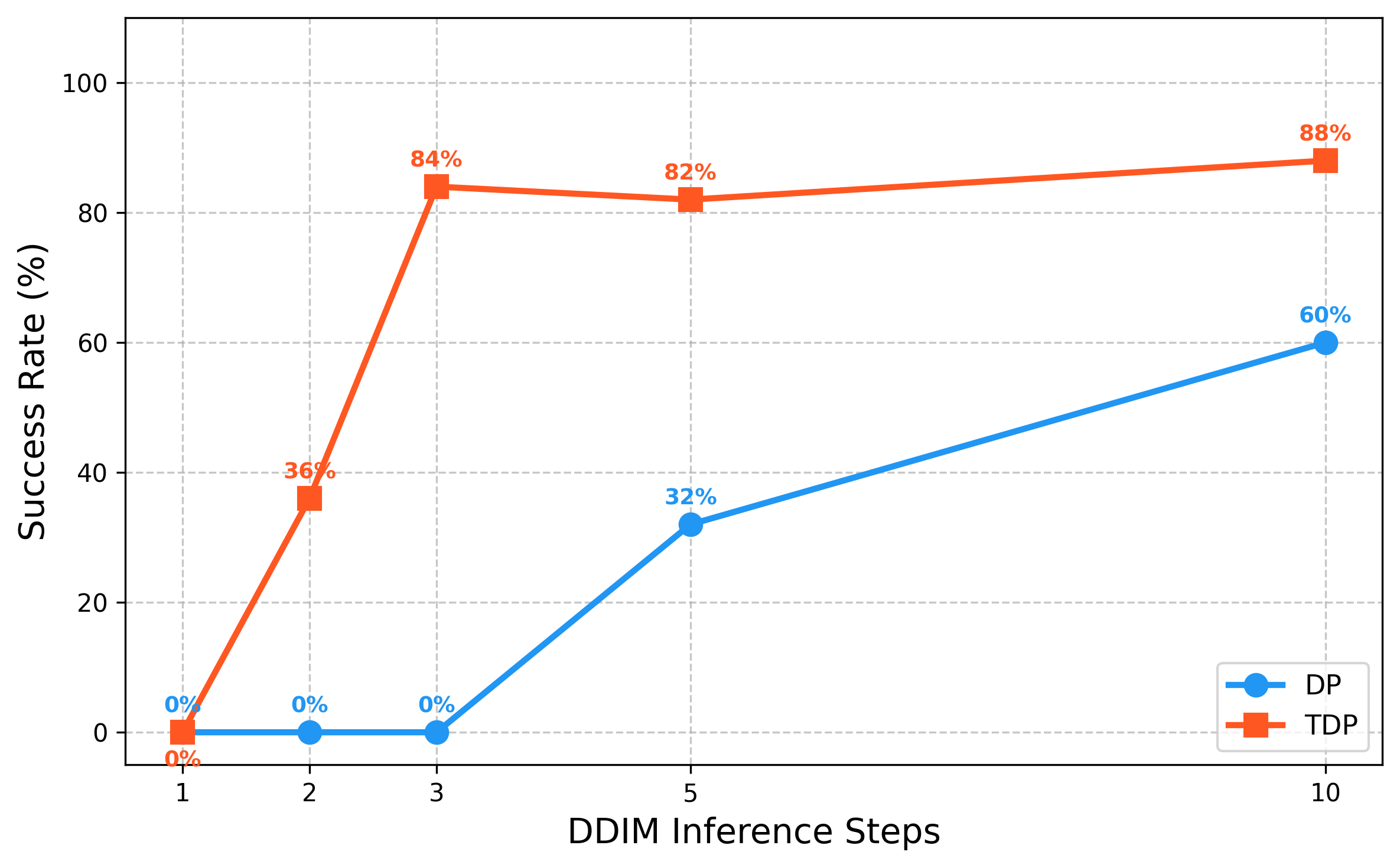}
        \caption{Stable grasping}
        \label{fig:ddim_ablation_reorientation}
    \end{subfigure}
    \caption{\textbf{Ablation study on DDIM inference steps.} We compare TDP and DP under varying numbers of DDIM inference steps on both grasping and stable grasping tasks. TDP consistently performs well with fewer inference steps than DP, demonstrating its ability to reduce control latency in practice. Furthermore, with the same number of inference steps, TDP achieves higher success rates owing to its step-wise reactive control mechanism.}
    \label{fig:ddim_ablation}
\end{figure*}

\begin{table}[t]
\centering
\setlength{\tabcolsep}{10pt}
\renewcommand{\arraystretch}{1.6}

\begin{tabular}{c c c c}
\toprule
\multirow{2}{*}{%
  \includegraphics[width=0.2\linewidth]{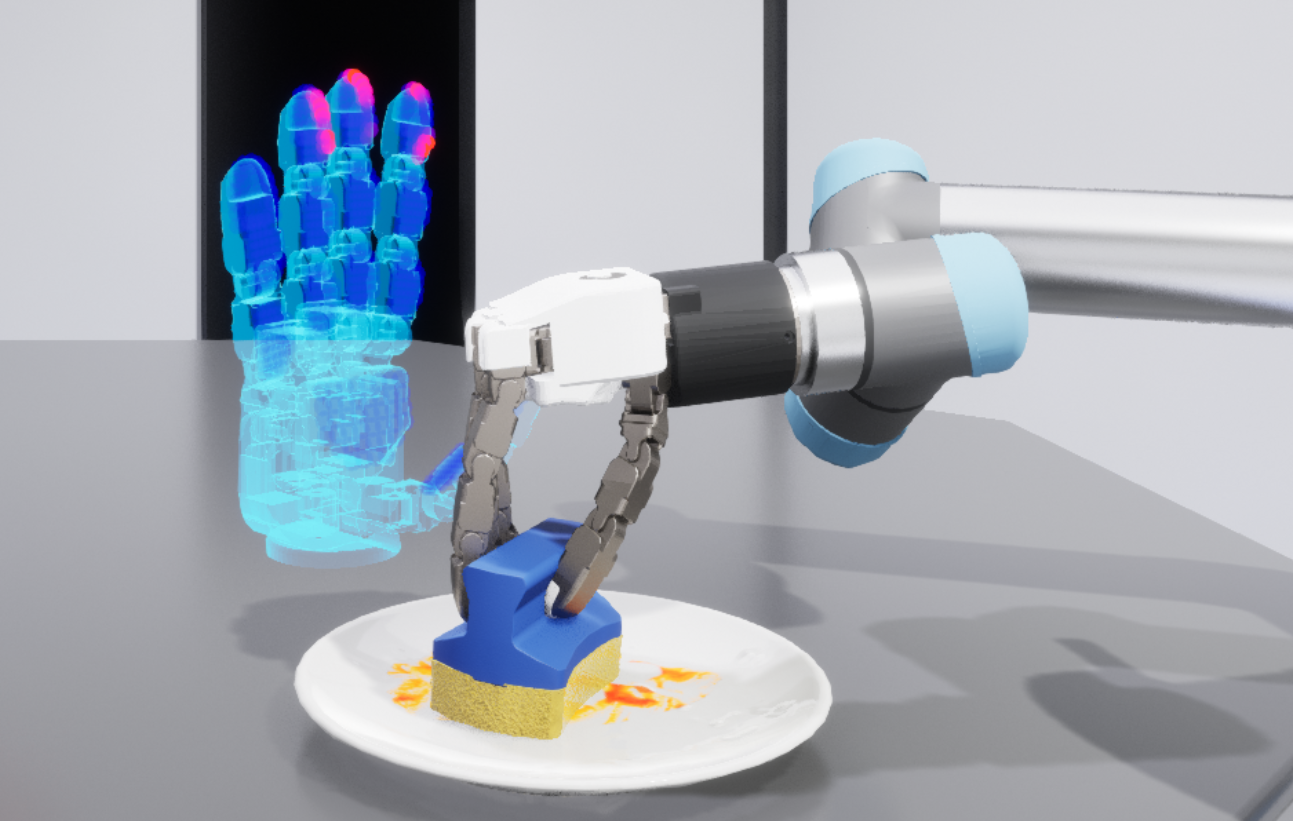}
}
 & \multirow{2}{*}{\textbf{Score}} 
 & \multicolumn{2}{c}{\textbf{Latency}} \\
\cmidrule(lr){3-4}

 &$\uparrow$   & Denoising ($\downarrow$) & Streaming ($\downarrow$) \\
\midrule

\rowcolor{orange!15}
DP (DDIM-10)
& \textbf{98.4\%} 
& \multicolumn{2}{c}{72.9 ms} \\

DP (DDIM-2)
& 5.3\%
& \multicolumn{2}{c}{17.1 ms} \\

\rowcolor{green!15}
\textbf{TDP} 
& 98.0\% 
& \textbf{13.3 ms} 
& \textbf{6.50 ms} \\

\bottomrule
\end{tabular}

\caption{\textbf{Comparison of DP and TDP on the dish-cleaning task.}  TDP achieves comparable success rate while significantly reducing inference latency.}
\label{tab:dish_comparison}
\end{table}

We evaluate our method on three additional visual–tactile dexterous manipulation tasks in a simulation environment built with Unreal Engine. The simulator integrates a real-time finite element solver through a native plugin, enabling physically realistic interactions between rigid and compliant objects. By leveraging reduced-order models, the solver efficiently simulates contact dynamics at interactive rates while providing detailed per-contact state information for dexterous manipulation \cite{tao2025interpolated, zhu2022viola, chang2023licrom, zong2023neural}. The simulated system consists of a UR5 robotic arm and a 20-DoF Tesollo DG5F dexterous hand equipped with a dense tactile array containing 768 sensing elements. Each element measures three-axis forces, resulting in a tactile observation of 2304 dimensions. Visual observations are obtained from two RGB cameras: one mounted above the workspace and another attached to the robot wrist. To collect demonstrations, we use a VR-based teleoperation interface with a Meta Quest 3 headset, as illustrated in Fig.~\ref{fig:teleopt_pipe}. The tracked human hand pose determines the pose of the robot end-effector, while the dexterous hand is controlled through fingertip retargeting. Specifically, the human fingertip positions are scaled to match the kinematic workspace of the DG5F hand. The corresponding finger joint targets are then computed using RelaxedIK \cite{rakita2018relaxedik, wang2023rangedik}. All simulation experiments are performed on a Lenovo P8 desktop equipped with an NVIDIA RTX 4080 GPU, responsible for robot control, visual–tactile sensing, and policy inference.

The first task, \emph{stable grasping}, requires the robot to grasp a mustard bottle and maintain a secure hold. Performance is evaluated using grasp success rate, stable grasp success rate, and steps to success, where a stable grasp requires holding the object for at least three seconds. The second task, \emph{on-table reorientation}, involves rotating a sugar box by 90 degrees via in-hand manipulation. The third task, \emph{dish cleaning}, requires cleaning a dirty dish using a sponge. Task performance is measured by the percentage of the dirty region cleaned. We collect 150 demonstrations for each task and evaluate each method over 50 episodes.

We first compare our approach against state-of-the-art baselines and conduct ablation studies on the \emph{stable grasping} and \emph{on-table reorientation} tasks, with results summarized in Table~\ref{tab:grasp_reorient_results}. Across both tasks, our method consistently outperforms all baselines in terms of success rate while requiring comparable or fewer steps to reach successful outcomes. In the \emph{stable grasping} task, our approach achieves a grasp success rate of 96\% and a stable grasp success rate of 88\%, substantially exceeding both DP and SFP, while maintaining efficient execution. Similar improvements are observed in the \emph{on-table reorientation} task, where our method attains the highest success rate (90\%) and reduces the number of steps compared to prior approaches. These results demonstrate the effectiveness of the proposed tube-based formulation in leveraging real-time visual–tactile feedback for robust and efficient manipulation. We further conduct an ablation study by removing the diffusion phase and relying solely on the streaming controller. In this setting, success rates drop significantly for both tasks. For \emph{on-table reorientation}, small errors tend to accumulate and often require multiple retries, leading to a higher number of steps to success despite eventual task completion. Interestingly, the streaming-only variant achieves the lowest number of steps in the \emph{stable grasping} task, which makes sense since this metric only accounts for completed episodes. In contrast to reorientation, failed grasp attempts typically cause the object to fall, making retries infeasible within the same episode. As a result, the steps-to-success values for grasping remain on a similar scale (and are even slightly better) than those of the full TDP method, as this metric mainly reflects the simplest successful scenarios. This further highlights the efficiency of step-wise reactive control for contact-rich manipulation, as this variant preserves a fully reactive formulation with fresh observations at every time step. 

Furthermore, we perform an ablation study on the number of DDIM inference steps to evaluate execution efficiency on the \emph{stable grasping} task (Fig.~\ref{fig:ddim_ablation}). With a single DDIM step, both TDP and DP fail to complete the task. As the number of inference steps increases, TDP exhibits a rapid performance gain with only two to three steps, whereas DP remains at a 0\% success rate and requires substantially more denoising steps (e.g., 5–10) to achieve similar performance. This indicates that the step-wise feedback regulation of the action tube can effectively compensate for coarse diffusion outputs, allowing TDP to attain high success rates with significantly fewer denoising steps and, consequently, reduced diffusion-stage inference latency for high-frequency, real-time control.

Based on this observation, we further evaluate both methods on the \emph{dish cleaning} task, with results shown in Table~\ref{tab:dish_comparison}. While DP with 10 DDIM steps achieves the highest cleaning score, it leads to significant control latency due to the long inference time. In contrast, TDP requires only 2 DDIM steps, making it possible to run at approximately 75~Hz during the denoising phase and up to 150~Hz during the streaming phase. When DP is constrained to a comparable inference budget (i.e., 2 DDIM steps), its performance drops significantly. This result highlights the capability of TDP for real-time, high-frequency control.

\subsection{Real-world experiments}

\begin{table*}[htbp]
\centering
\setlength{\tabcolsep}{10pt}
\renewcommand{\arraystretch}{1.6}

\begin{tabular}{c c c c c}
\toprule
\multirow{2}{*}{%
  \includegraphics[width=0.1\linewidth]{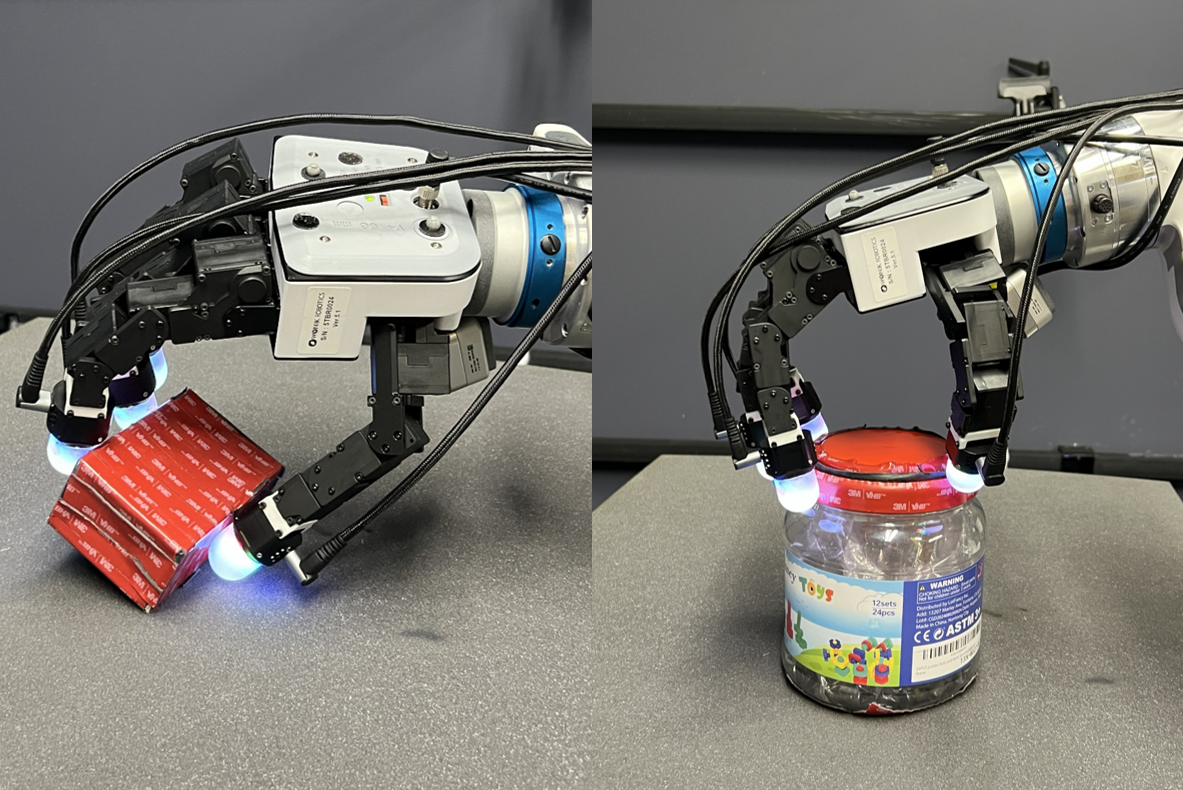}
}
 & \multirow{2}{*}{\textbf{On-table Reorientation}} 
 & \multirow{2}{*}{\textbf{Jar Opening}} 
 & \multicolumn{2}{c}{\textbf{Latency}} \\
\cmidrule(lr){4-5}

 &Success rate ($\uparrow)$  &Success rate ($\uparrow$)   & Denoising ($\downarrow$) & Streaming ($\downarrow$) \\
\midrule

\rowcolor{orange!15}
DP
& 60\% & 84\% 
& \multicolumn{2}{c}{0.037s} \\

\rowcolor{green!15}
\textbf{TDP} 
& \textbf{96\%} & \textbf{96\%} 
& \textbf{0.008s} 
& \textbf{0.003s} \\

\bottomrule
\end{tabular}

\caption{\textbf{Comparison of DP and TDP on Two Physical Experiments.}  TDP achieves higher success rate with significantly lower inference latency.}
\label{tab:quant_phyx_comparison}
\end{table*}

\begin{figure*}[htbp]
\centering
    \includegraphics[width=0.9\linewidth]{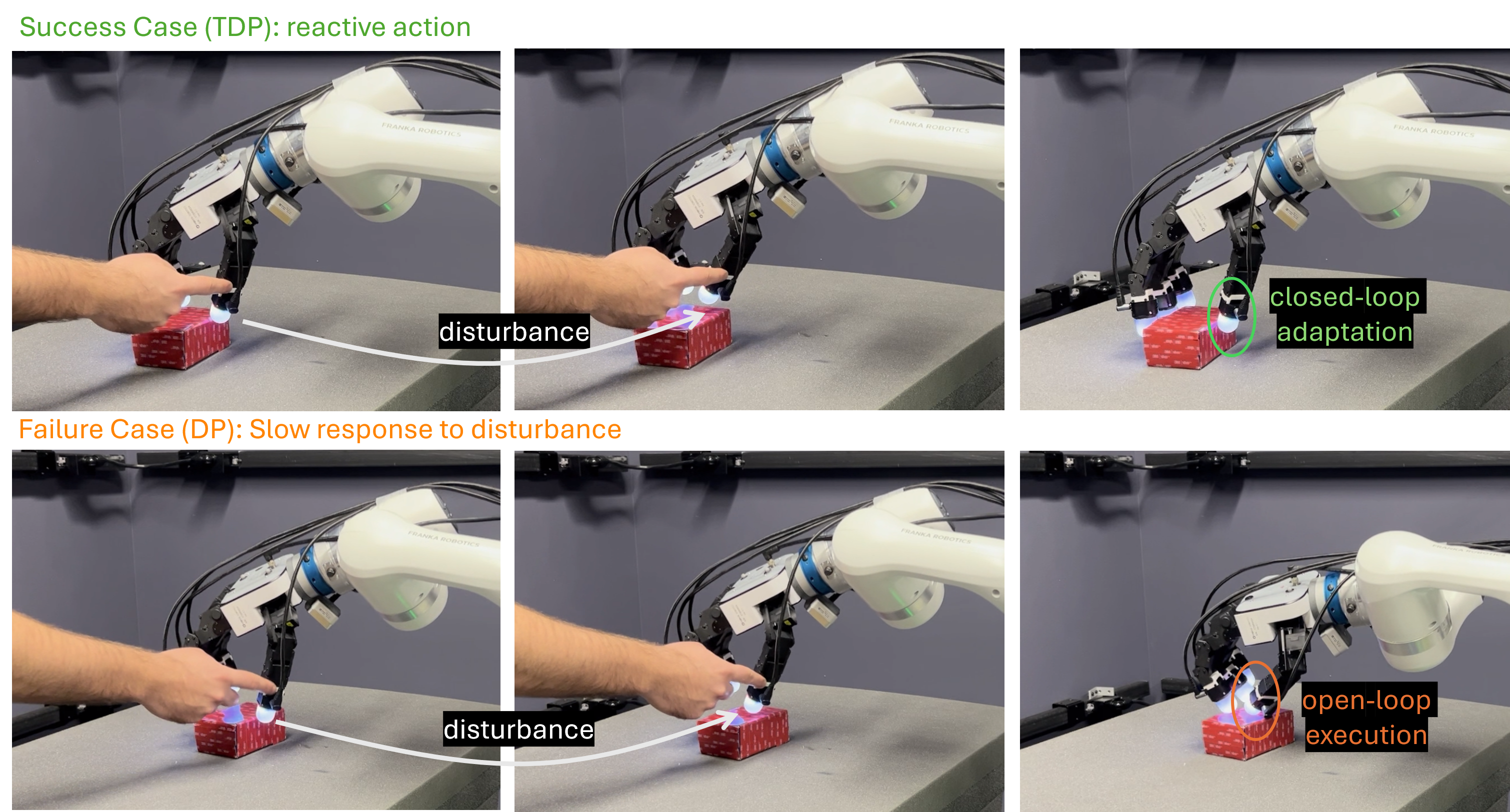}
    \caption{\textbf{On-table manipulation task under external disturbance}. The disturbance is introduced by bending the thumb fingers of the Allegro hand. Tube Diffusion Policy (TDP) reacts quickly to the perturbation, while Diffusion Policy (DP) exhibits a delayed response and the fingers become stuck, resulting in task failure.}
    \label{fig:flip_adapt}
\end{figure*}

\begin{figure*}[htbp]
\centering
    \includegraphics[width=0.9\linewidth]{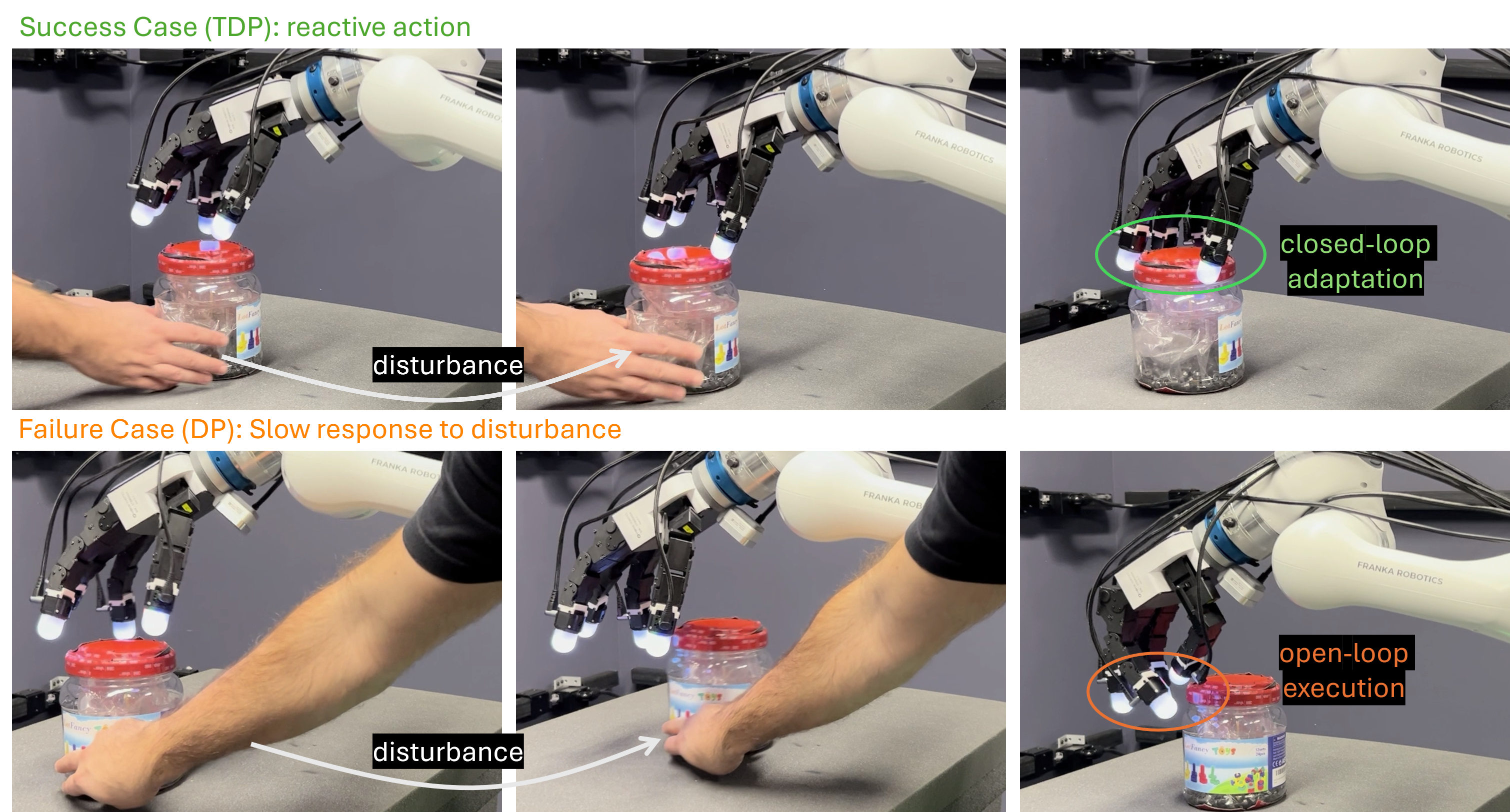}
    \caption{\textbf{Jar-opening task under external disturbance}. We introduce disturbance by shifting the jar’s position. Tube Diffusion Policy (TDP) quickly adapts to the new position and continues opening the jar, whereas Diffusion Policy (DP) blindly executes the pre-generated action sequence, resulting in slow response to the disturbance.}
    \label{fig:jar_adapt}
\end{figure*}

\begin{figure*}[htbp]
\centering
\begin{subfigure}[t]{0.48\linewidth}
    \centering
    \includegraphics[width=\linewidth]{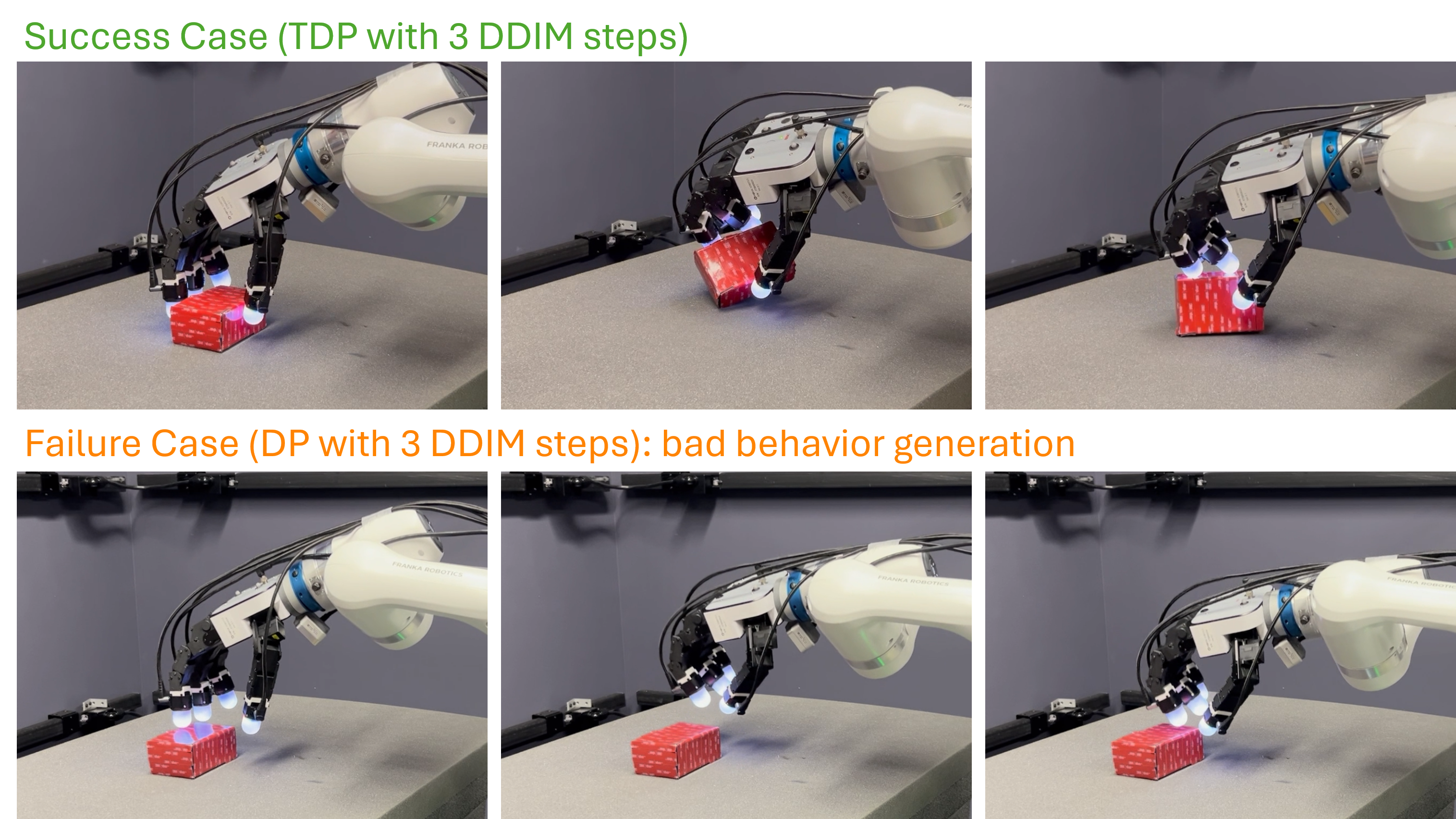}
    \caption{On-table reorientation task with 3 DDIM steps.}
\end{subfigure}
\hfill
\begin{subfigure}[t]{0.48\linewidth}
    \centering
    \includegraphics[width=\linewidth]{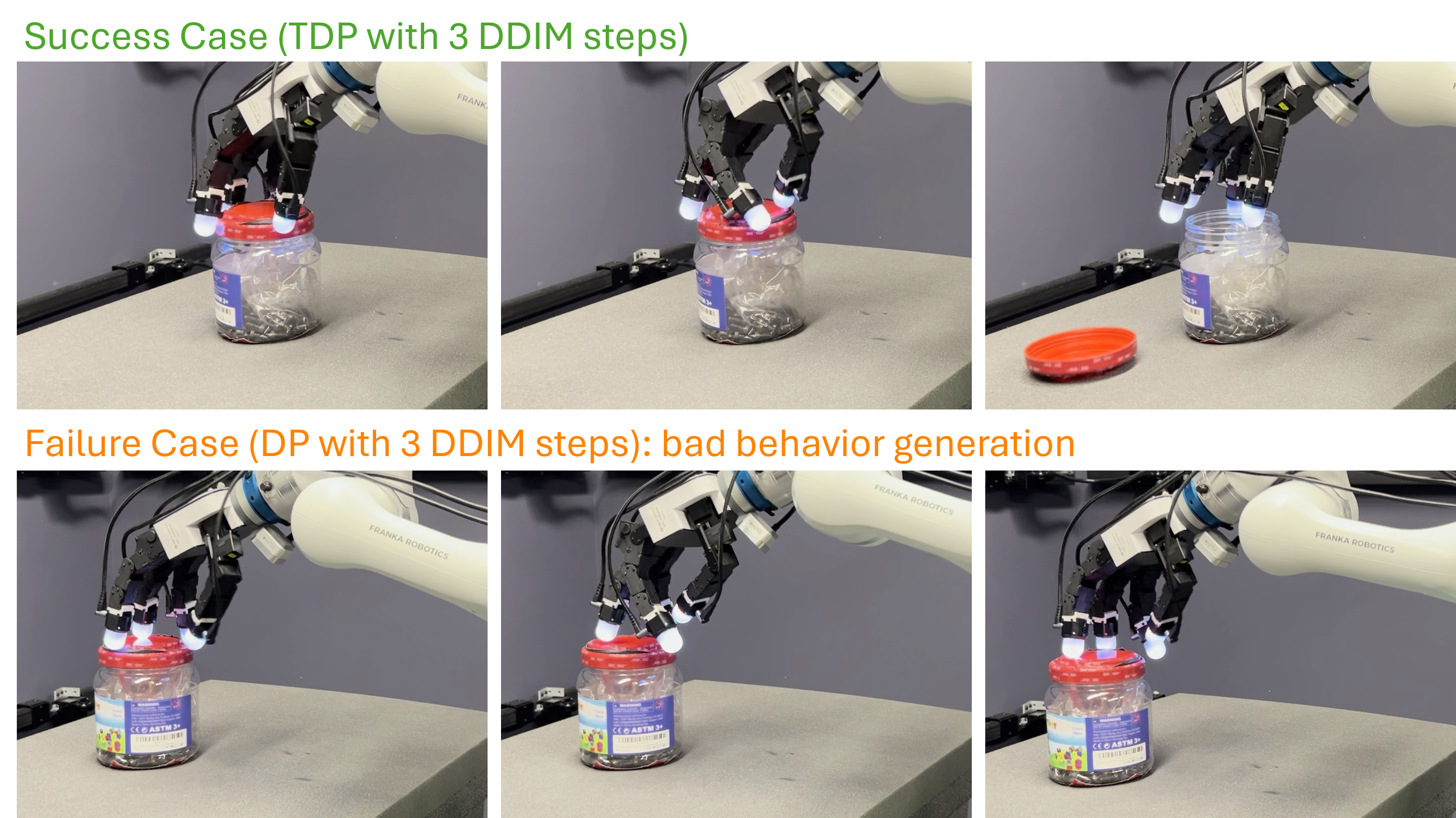}
    \caption{Jar-opening task with 3 DDIM steps.}
\end{subfigure}

\caption{\textbf{Real-world manipulation with few DDIM inference steps.} 
Tube Diffusion Policy (TDP) maintains strong performance with only three DDIM inference steps, enabling high-frequency control in the real world. In contrast, Diffusion Policy (DP) produces inaccurate behaviors under the same inference budget, indicating its reliance on more DDIM steps for stable execution and consequently higher control latency.}
\label{fig:ddim3_real}
\end{figure*}

We further validate TDP on two real-world visual–tactile dexterous manipulation tasks: jar opening and on-table reorientation, as illustrated in Fig.~\ref{fig:firstpage}. These tasks involve contact-rich interactions, pose uncertainty, and frequent variations in hand–object contact, posing significant challenges for policies relying on open-loop execution.

All physical experiments are conducted using a Franka Research 3 robotic arm equipped with an Allegro V5 dexterous hand. Visual observations are provided by an Intel RealSense D405 camera for the wrist-mounted view and a D435F camera for the global view. In addition, Digit360 tactile sensors are attached to the fingertips to capture tactile feedback during manipulation. The system runs on two workstations. A Lenovo P620 desktop (NVIDIA RTX 3080 GPU) is used for low-level robot control and real-time visual–tactile sensing processing, while a Lenovo P8 desktop (NVIDIA RTX 4080 GPU) is dedicated to policy inference. To collect demonstrations, we employ a teleoperation system based on an OptiTrack motion-capture setup together with an instrumented glove for hand tracking, as illustrated in Fig.~\ref{fig:teleopt_pipe}. The pose of the robot end-effector is determined by tracking the pose of the human hand. For hand teleoperation, we adopt a fingertip retargeting strategy: fingertip positions estimated from the glove are mapped to the workspace of the Allegro hand while ignoring the human pinky finger. We then solve the inverse kinematics using RelaxedIK \citep{rakita2018relaxedik, wang2023rangedik} to obtain the desired finger joint targets.

We first conduct 25 evaluation episodes for each task using both TDP and DP. The quantitative results are summarized in Table \ref{tab:quant_phyx_comparison}. Benefiting from its closed-loop reactive control, TDP achieves significantly higher success rates on both tasks. Moreover, TDP allows the use of fewer DDIM inference steps thanks to the tube-based correction mechanism applied at each timestep. In our experiments, we use only 3 DDIM steps during denoising, which results in substantially lower inference latency and enables control frequencies above 100 Hz, approximately four times faster than DP ($\sim$ 25 Hz).

To further examine this property, we also evaluate DP using the same number of DDIM inference steps for high-frequency control. As shown in Fig.~\ref{fig:ddim3_real}, DP with only 3 DDIM steps produces inaccurate behaviors and fails to complete the tasks, indicating that DP requires more inference steps for stable action generation.

We additionally evaluate the step-wise reactive capability of TDP by introducing external disturbances during task execution. Fig.~\ref{fig:flip_adapt} and Fig.~\ref{fig:jar_adapt} show representative keyframes before and after disturbances. In the on-table reorientation task, we introduce perturbations by bending the thumb finger of the Allegro hand. As shown in Fig.~\ref{fig:flip_adapt}, TDP quickly reacts to the perturbation and recovers to a stable configuration. In contrast, DP continues executing the pre-generated action sequence, causing the fingers to become stuck on the object and resulting in task failure. In the jar-opening task, we introduce disturbance by shifting the position of the jar during execution. As shown in Fig.~\ref{fig:jar_adapt}, TDP rapidly tracks the updated jar pose and continues the opening motion. In contrast, DP blindly executes the pre-generated action chunk, failing to adapt to the changed object position.

Overall, the physical experiments demonstrate that TDP consistently achieves higher success rates and more stable manipulation behaviors than DP. When unexpected contact changes or disturbances occur, TDP can react immediately and adjust the hand configuration online. In contrast, DP tends to continue executing pre-generated action sequences in an open-loop manner, which often leads to failure.

Importantly, TDP supports inference with very few DDIM steps, significantly reducing latency and enabling fast, high-frequency control. Combined with its per-step reactive control within the learned action tube, this makes TDP particularly well suited for real-world contact-rich dexterous manipulation.

\section{Limitations and Future Work}
\label{sec:limitation}

While we have demonstrated the effectiveness of our approach in both simulation and real-world settings, several limitations remain and point to promising directions for future work. 
First, we simply concatenate visual, tactile and proprioceptive perception as multi-modal observation, which neglects the inherent sparsity and asynchronous nature of tactile sensing. Future work could explore more structured or learned tactile representations to better exploit contact information. Second, although tube-based formulation allows the use of a small number of DDIM inference steps to reduce latency, further acceleration could be achieved by leveraging recent advances in diffusion model acceleration techniques, such as one-step or distillation-based denoising methods \cite{song2023consistency, prasad2024consistency, frans2025one}. Third, while this work focuses on improving standard diffusion policies, the proposed tube-based formulation is broadly applicable to many existing chunk-based or open-loop policy learning methods. Extending this framework to other generative or sequence-based control approaches remains an important direction for future research. For example, most Vision–Language–Action (VLA) models \cite{black2024pi_0, intelligence2025pi_, black2025real} rely on flow-based action chunking, which could benefit from integrating the \emph{action tube} formulation to enable fast and reactive execution. Overall, we believe this approach provides a general pathway for mitigating the limitations of open-loop chunk execution and improving robustness in contact-rich manipulation.

\section{Conclusion} 
\label{sec:conclusion}
In this work, we present \emph{Tube Diffusion Policy} (TDP), a reactive visual-tactile policy learning framework that augments diffusion-based imitation learning with feedback regulation during action execution. In contrast to conventional action chunking approaches that execute pre-generated action sequences in an open-loop manner, TDP constructs an \emph{action tube} that enables step-wise feedback control conditioned on fresh observations. This structure allows the policy to react to unexpected contacts and external disturbances during execution, which is particularly important for contact-rich dexterous manipulation.

We evaluate TDP on three visual–tactile manipulation tasks in simulation and two tasks in the real world. Experimental results demonstrate that the proposed approach consistently improves task success rates and robustness compared with state-of-the-art imitation learning methods. In real-world experiments, TDP further demonstrates rapid recovery from external disturbances and unexpected contact variations during execution. Moreover, TDP does not require highly accurate generation of full action sequences. Instead, the diffusion phase only needs to produce a coarse corrective action that accounts for the full system dynamics, after which actions are generated by streaming conditional feedback flows. This design substantially reduces inference latency and enables high-frequency control in practice.

Overall, this work provides a principled bridge between generative policy learning and feedback control, addressing a key limitation of existing chunk-based methods—open-loop execution under uncertainty. We believe the proposed tube-based formulation offers a promising direction for enabling robust, reactive, and high-frequency control in visual-tactile contact-rich manipulation.

\bibliographystyle{plainnat}
\bibliography{references}

\clearpage
\appendix
\subsection{Stability Analysis of Tube Diffusion Policy}

This section provides a more detailed analysis and formal proof of the stability properties of Tube Diffusion Policy (TDP). TDP decomposes control into two complementary mechanisms: diffusion correction, which periodically contracts the accumulated drift, and streaming imitation, which provides high-frequency reactive control but introduces bounded drift. Together, these mechanisms form a hybrid control system that admits a practical stability guarantee.



\subsubsection{Problem Setup}

Consider the discrete-time nonlinear system
\begin{equation}
\bm{o}_{t+1} = g(\bm{o}_t,\bm{u}_t) + \bm{w}_t,
\label{eq:system}
\end{equation}
where \(\bm{o}_t \in \mathbb{R}^{n_o}\) is the system state, \(\bm{u}_t \in \mathbb{R}^{n_u}\) is the control input, and \(\bm{w}_t\) is a bounded disturbance.

We assume access to a reference demonstration trajectory
\begin{equation}
(\bm{o}_t^*, \bm{u}_t^*)
\end{equation}
that satisfies the nominal dynamics
\begin{equation}
\bm{o}_{t+1}^* = g(\bm{o}_t^*, \bm{u}_t^*).
\label{eq:reference_dyn}
\end{equation}

Tube Diffusion Policy consists of two components:

\paragraph{Diffusion correction}
At correction instants \(t_k\), a diffusion model produces a corrective action that compensates for the drift accumulated during streaming execution.

\paragraph{Streaming flow policy}
A learned policy generates actions sequentially,
\begin{equation}
\bm{u}_{t+1} = \bm{u}_t + v_\theta(\bm{u}, t_1 = 0, t_2 \,|\, h) \cdot \Delta t,
\end{equation}
and is trained to imitate the demonstration action trajectory.

We define the tracking error
\begin{equation}
\bm{e}_t := \bm{o}_t - \bm{o}_t^*.
\label{eq:error_def}
\end{equation}

\subsubsection{Assumptions}

To analyze the stability of Tube Diffusion Policy, we introduce the following assumptions. 
Assumptions~1 and~3 characterize properties of the system dynamics and disturbances, while Assumptions~2 and~4 describe the desired behavior of the learned streaming and diffusion policies induced by training.

\textbf{Assumption 1 (Lipschitz dynamics).}
There exist constants \(L_x > 0\) and \(L_u > 0\) such that for all admissible states and actions,
\begin{equation}
\|g(\bm{o}_1,\bm{u}_1) - g(\bm{o}_2,\bm{u}_2)\|
\le
L_x \|\bm{o}_1-\bm{o}_2\| + L_u \|\bm{u}_1-\bm{u}_2\|.
\label{eq:lipschitz}
\end{equation}

The Lipschitz condition bounds the sensitivity of the system dynamics to state and action deviations, enabling tractable propagation of tracking errors.

\textbf{Assumption 2 (Streaming imitation accuracy).}
During streaming execution, the learned policy approximately imitates the reference actions, resulting in a bounded action error:
\begin{equation}
\|\bm{u}_t - \bm{u}_t^*\| \le \varepsilon_a.
\label{eq:imitation_error}
\end{equation}

This condition reflects the training objective of the streaming phase and is expected to hold when the policy accurately reproduces the demonstrated actions.

\textbf{Assumption 3 (Bounded disturbance).}
The external disturbance is bounded:
\begin{equation}
\|\bm{w}_t\| \le \bar w.
\label{eq:disturbance_bound}
\end{equation}

\textbf{Assumption 4 (Diffusion correction property).}
At each correction instant \(t_k\), the diffusion module produces a corrective action that contracts the tracking error:
\begin{equation}
\|\bm{e}_{t_k+1}\| \le \lambda \|\bm{e}_{t_k}\| + \varepsilon_d,
\qquad 0 \le \lambda < 1.
\label{eq:diffusion_contraction}
\end{equation}

This condition corresponds to the training objective of the diffusion phase, which seeks to reduce the accumulated tracking error at correction steps by accounting for the full system dynamics.


\subsubsection{Streaming Phase Error Dynamics}

From \eqref{eq:system}, \eqref{eq:reference_dyn}, and \eqref{eq:error_def}, the error dynamics during the streaming phase satisfy
\begin{align}
\bm{e}_{t+1}
&= g(\bm{o}_t,\bm{u}_t) + \bm{w}_t - g(\bm{o}_t^*,\bm{u}_t^*) \\
&= g(\bm{o}_t,\bm{u}_t) - g(\bm{o}_t^*,\bm{u}_t^*) + \bm{w}_t.
\end{align}

Applying Assumption 1 gives
\begin{equation}
\|\bm{e}_{t+1}\|
\le
L_x \|\bm{e}_t\| + L_u \|\bm{u}_t-\bm{u}_t^*\| + \|\bm{w}_t\|.
\end{equation}

Using Assumptions 2 and 3,
\begin{equation}
\|\bm{e}_{t+1}\|
\le
L_x \|\bm{e}_t\| + c,
\qquad
c := L_u \varepsilon_a + \bar w.
\label{eq:error_recursion}
\end{equation}

\subsubsection{Lyapunov Function for the Error Dynamics}

Consider the Lyapunov candidate
\begin{equation}
V(\bm{e}_t) := \|\bm{e}_t\|.
\label{eq:lyapunov_candidate}
\end{equation}
This function is positive definite and radially unbounded with respect to the tracking error \(\bm e_t\). Then \eqref{eq:error_recursion} implies
\begin{equation}
V(\bm{e}_{t+1}) \le L_x V(\bm{e}_t) + c.
\label{eq:V_stream}
\end{equation}
Thus, the streaming phase is input-to-state stable with respect to the combined input \(c\), although it does not necessarily contract to zero in the absence of diffusion correction.

\begin{lemma}[Streaming-phase ISS bound]
\label{lem:streaming_iss}
For any integer \(j \ge 0\), the tracking error during the streaming phase satisfies
\begin{equation}
V(\bm{e}_{t+j})
\le
L_x^j V(\bm{e}_t)
+
\sum_{i=0}^{j-1} L_x^i c.
\label{eq:streaming_unrolled}
\end{equation}
If \(L_x < 1\), then
\begin{equation}
V(\bm{e}_{t+j})
\le
L_x^j V(\bm{e}_t)
+
\frac{1-L_x^j}{1-L_x} c.
\label{eq:streaming_geo}
\end{equation}
\end{lemma}

\begin{proof}
The result follows by recursively unrolling \eqref{eq:V_stream}. When \(L_x<1\), the geometric series identity
\[
\sum_{i=0}^{j-1}L_x^i = \frac{1-L_x^j}{1-L_x}
\]
yields \eqref{eq:streaming_geo}.
\end{proof}

\subsubsection{Error Evolution Across Correction Cycles}

The diffusion correction is applied every \(H_a\) steps.
Let \(z_k := V(\bm{e}_{t_k})\) denote the tracking error immediately
after the diffusion correction at time \(t_k\).

After the correction at \(t_k\), the system evolves under the
streaming controller for the next \(H_a-1\) steps,
from \(t_k+1\) to \(t_k+H_a-1\).
By Lemma~\ref{lem:streaming_iss}, the error at time \(t_k+H_a-1\)
satisfies
\begin{equation}
V(\bm{e}_{t_k+H_a-1})
\le
L_x^{H_a-1} z_k
+
\frac{1-L_x^{H_a-1}}{1-L_x} c .
\label{eq:streaming_growth}
\end{equation}

At time \(t_{k+1}=t_k+H_a\), a new diffusion correction is applied.
Using Assumption~4 gives
\begin{equation}
z_{k+1}
\le
\lambda
\left(
L_x^{H_a-1} z_k
+
\frac{1-L_x^{H_a-1}}{1-L_x} c
\right)
+
\varepsilon_d .
\label{eq:zk_recursion}
\end{equation}

Equivalently,
\begin{equation}
z_{k+1}
\le
\alpha z_k + \beta,
\label{eq:linear_recursion}
\end{equation}
where
\begin{equation}
\alpha := \lambda L_x^{H_a-1},
\qquad
\beta := \lambda \frac{1-L_x^{H_a-1}}{1-L_x} c + \varepsilon_d .
\label{eq:alpha_beta}
\end{equation}

\begin{lemma}[Contraction across correction cycles]
\label{lem:cycle_contraction}
Suppose \(L_x<1\). Then \(0 \le \alpha < 1\), and the post-correction
error sequence satisfies
\begin{equation}
z_k
\le
\alpha^k z_0 + \frac{1-\alpha^k}{1-\alpha}\beta.
\label{eq:zk_solution}
\end{equation}
Hence,
\begin{equation}
\limsup_{k\to\infty} z_k \le \frac{\beta}{1-\alpha}.
\label{eq:zk_limsup}
\end{equation}
\end{lemma}

\begin{proof}
Since \(0 \le \lambda < 1\) and \(L_x < 1\), we have
\(0 \le \alpha = \lambda L_x^{H_a-1} < 1\).
The result follows by unrolling the affine recursion
\eqref{eq:linear_recursion}.
\end{proof}

\subsubsection{Main Result}

\begin{theorem}[Practical stability of Tube Diffusion Policy]
\label{thm:tdp_practical_stability}
Suppose Assumptions 1--4 hold and \(L_x < 1\). Then the closed-loop system under TDP satisfies the following properties:

\textbf{(1) Uniform ultimate boundedness.}
The tracking error sequence is uniformly ultimately bounded, i.e.,
\[
\limsup_{t\to\infty}\|\bm e_t\| < \infty .
\]

\textbf{(2) Bounded error within each correction cycle.}
For any correction cycle \(t \in \{t_k,\dots,t_k+H_a-1\}\), the tracking error remains bounded as a function of the post-correction error \(z_k\).
\end{theorem}

\begin{proof}
First, by Lemma~\ref{lem:cycle_contraction}, the post-correction error satisfies the affine recursion
\[
z_{k+1} \le \alpha z_k + \beta ,
\]
with \(0 \le \alpha < 1\). Standard results for linear difference inequalities imply
\[
\limsup_{k\to\infty} z_k \le \frac{\beta}{1-\alpha}.
\]
Hence the sequence of post-correction errors is uniformly ultimately bounded.

Next, consider any time \(t \in \{t_k,\dots,t_k+H_a-1\}\). By Lemma~\ref{lem:streaming_iss},
\[
\|\bm e_t\|
\le
L_x^{H_a-1} z_k + \frac{1-L_x^{H_a-1}}{1-L_x}c .
\]
Thus the tracking error remains bounded during each streaming phase as a function of \(z_k\). Combining this result with the boundedness of \(z_k\) establishes uniform ultimate boundedness of the full error trajectory.

In the ideal case where \(\varepsilon_a=0\), \(\bar w=0\), and \(\varepsilon_d=0\), we have \(c=0\) and \(\beta=0\), so the recursion reduces to
\[
z_{k+1} \le \alpha z_k ,
\qquad 0 \le \alpha < 1 .
\]
Hence \(z_k \to 0\). Using Lemma~\ref{lem:streaming_iss} with \(c=0\) further implies \(\bm e_t \to 0\), and the closed-loop system is asymptotically stable.

\end{proof}


In summary, Theorem~\ref{thm:tdp_practical_stability} formalizes the role of the two components in TDP. 
At each correction cycle, the diffusion module first performs a correction step, followed by \(H_a-1\) streaming control steps. The streaming flow provides high-frequency reactive control. However, due to bounded imitation error and external disturbances, it introduces a bounded tracking drift during the streaming phase. The diffusion module periodically contracts this accumulated drift at the beginning of each correction cycle. Their combination yields a hybrid closed-loop system in which the tracking error remains within a bounded tube around the demonstrated reference trajectory.
This matches the intuition behind TDP: streaming flow ensures responsive step-wise control, while diffusion correction prevents long-term drift accumulation.

\subsection{Network Architecture and Hyperparameters}
\label{app:architecture}

\begin{table}[t]
\caption{Network Architecture and Hyperparameters}
\label{tab:architecture}
\centering
\small
\begin{tabular}{p{3.2cm} p{4.8cm}}
\hline
\textbf{Component} & \textbf{Specification} \\
\hline
U-Net backbone 
& 1D U-Net with 3 resolution levels (channels 256, 512, 1024) \\

Residual blocks 
& 12 FiLM-conditioned residual blocks \\

Backbone encoders 
& 6 ResNet-18 encoders (2 vision, 4 tactile; no weight sharing) \\

FiLM conditioning 
& Linear layer with conditioning dimension 3118 generating scale and shift \\

U-Net backbone 
& 44.7M \\

Policy network 
& 105.2M \\

Full model 
& 172.8M \\
\hline
\end{tabular}
\end{table}

We provide additional details of the policy network architecture. 
Our method employs a conditional 1D U-Net backbone for diffusion-based action generation 
and streaming flow, augmented with visual and tactile encoders.

\paragraph{U-Net backbone.}
The policy network is a 1D U-Net with three resolution levels 
(channel dimensions 256, 512, and 1024), consisting of symmetric downsampling and 
upsampling stages with 12 FiLM-conditioned residual blocks.

\paragraph{Conditioning.}
Observation conditioning is incorporated via feature-wise linear modulation (FiLM). 
The conditioning dimension is 3118, and FiLM layers produce per-channel scale and shift parameters.

\paragraph{Observation encoders.}
Visual and tactile inputs are encoded using six ResNet-18 backbones 
(two for vision and four for tactile inputs, without parameter sharing).

\paragraph{Model size.}
The policy U-Net contains 44.7M parameters. 
The full policy network, including the U-Net backbone and action prediction heads but excluding observation encoders, contains 105.2M parameters. 
The complete model, including all visual and tactile encoders, contains 172.8M parameters.

Table~\ref{tab:architecture} summarizes the key components and hyperparameters.

\end{document}